\documentclass[runningheads]{llncs}
\usepackage[T1]{fontenc}
\usepackage{graphicx}
\usepackage{booktabs}
\usepackage[misc]{ifsym}
\newcommand{\corr}{(\Letter)}
\usepackage[margin=1in]{geometry}
\usepackage{cite}
\usepackage{amsmath,amssymb,amsfonts}
\usepackage{mathtools}
\usepackage{algorithmic}
\usepackage{graphicx}
\usepackage{textcomp}
\usepackage{xcolor}
\usepackage{bbm, bm}
\usepackage{xcolor}
\usepackage{booktabs}
\usepackage{enumitem} 
\usepackage{cuted}
\usepackage{comment}
\usepackage{hyperref}

\newcommand{\bs}[1]{\boldsymbol{#1}}
\def\BibTeX{{\rm B\kern-.05em{\sc i\kern-.025em b}\kern-.08em
    T\kern-.1667em\lower.7ex\hbox{E}\kern-.125emX}}
\begin{document}

\title{Identifiable Autoregressive Variational Autoencoders for Nonlinear and Nonstationary Spatio-Temporal Blind Source Separation
\thanks{This research was supported by the Research Council of Finland (363261, 453691) and the Vilho, Yrjö and Kalle Väisälä foundation.}
}
\titlerunning{Identifiable Autoregressive Variational Autoencoders}

\author{Mika Sipilä \\
	Department of Mathematics and Statistics\\
	University of Jyvaskyla\\
	Finland \\
        \And
	Klaus Nordhausen \\
	Department of Mathematics and Statistics\\
	University of Jyvaskyla\\
	Finland \\
        \And
        Sara Taskinen \\
	Department of Mathematics and Statistics\\
	University of Jyvaskyla\\
	Finland \\
}

\author{Mika Sipilä\inst{1}\corr \and
Klaus Nordhausen\inst{2} \and
Sara Taskinen\inst{1}}

\authorrunning{M. Sipilä et al.}

\institute{Department of Mathematics and Statistics, University of Jyvaskyla, Finland
\and
Department of Mathematics and Statistics, University of Helsinki, Finland }

\maketitle              

\begin{abstract}
The modeling and prediction of multivariate spatio-temporal data involve numerous challenges. Dimension reduction methods can significantly simplify this process, provided that they account for the complex dependencies between variables and across time and space. Nonlinear blind source separation has emerged as a promising approach, particularly following recent advances in identifiability results. Building on these developments, we introduce the identifiable autoregressive variational autoencoder, which ensures the identifiability of latent components consisting of nonstationary autoregressive processes. The blind source separation efficacy of the proposed method is showcased through a simulation study, where it is compared against state-of-the-art methods, and the spatio-temporal prediction performance is evaluated against several competitors on air pollution and weather datasets. 
\end{abstract}

\section{Introduction}

In multivariate spatio-temporal data, the multivariate observations $\bs x(\bs s, t) \coloneqq \bs x^t \coloneqq \bs x \in \mathcal{X} \subset \mathbb{R}^S$ are collected in various spatial locations $\bs s \in \mathcal{S} \subset \mathbb{R}^D$ at times $t \in \mathcal{T} \subset \mathbb{R}$, where $\mathcal{X}$ is the domain of $\bs x$, $\mathcal{S}$ and $\mathcal{T}$ are spatial and temporal domains, respectively, and $D$ is a spatial dimension. Modeling and predicting such data are highly challenging and computationally demanding due to the fact that the spatio-temporal dependency structures, as well as the dependencies between the variables, have to be accounted for. These dependencies are often modeled through $S \times S$ dimensional covariance function $\bs C(\bs x(\bs s, t), \bs x(\bs s', t'))$. Modeling the covariance function is especially complicated in case of nonstationary 
data \cite{salvana2020nonstationary, PorcuFurrerNychka2021}, which means that the covariance function $\bs C$ cannot be simplified to stationary form $\bs C(\bs x(\bs s, t), \bs x(\bs s', t')) = \bs C(\|\bs s - \bs s'\|, |t - t'|)$. Instead, for nonstationary data, the covariance function $\bs C$ changes when spatial or temporal locations are shifted.


Spatio-temporal data modeling can be simplified without restrictive assumptions like stationarity, by using blind source separation. In blind source separation, it is assumed that the observation $\bs x$ is generated from the independent latent component $\bs z(\bs s, t) \coloneqq \bs z^t \coloneqq \bs z \in \mathbb{R}^P$ through a mixing function $\bs f$ as 
\begin{align}
    \bs x = \bs f(\bs z).
    \label{eq:nonlinearBSS}
\end{align} 

Once the latent components are successfully recovered, they can be modeled independently due to their assumed statistical independence. The dependencies among the components of the observed variable vector $\bs{x}$ are therefore presumed to arise exclusively from the mixing function $\bs{f}$. Blind source separation (BSS) aims to recover the latent components by estimating the mixing and unmixing functions from the observed data.

While most traditional BSS methods, such as spatio-temporal BSS (STBSS) \cite{MuehlmannDeIacoNordhausen2022}, are limited only to linear mixing function $\bs f(\bs z) = \bs A\bs z$, where $\bs A$ is a $S \times P$ matrix, nonlinear BSS variants have also been recently developed. In the nonlinear case however stronger assumptions are needed for identifiability. One such approach for nonlinear BSS assumes, for example, structural sparsity \cite{lachapelle2022disentanglement}. Other recent developments are mostly for time series, and they solve nonlinear BSS by exploiting either stationary autocorrelation structure or nonstationary variances. For these methods, see \cite{hyvarinen2023nonlinear} and the references therein.

In particular, \cite{Khemakhem2020} introduced identifiable variational autoencoder (iVAE) for nonlinear and nonstationary temporal BSS. Later, iVAE have been extended to nonstationary spatial data in \cite{SipilaNonlinearSBSS} and to nonstationary spatio-temporal data in \cite{sipila2024modelling}. However, all previous iVAE methods are identifiable only if the latent components possess nonstationary variance, and they do not incorporate previous observations in time in the model. Instead, the previous methods model the nonstationary variance only based on the spatial and temporal location of the observations.

In this paper, we assume that each latent component $z_i$, for $i = 1, \dots, P$, is generated by a nonstationary autoregressive process defined as follows:
\begin{align}
    z_i(\bs s, t) = \mu_i(\bs s, t) + \sum_{r=1}^R \gamma_{i, r}(\bs s, t) \Bigl( z_i(\bs s, t-r) - \mu_i(\bs s, t-r) \Bigr) + \omega_i(\bs s, t),
    \label{eq:ar_process}
\end{align}
where $\mu_i$ is a nonstationary trend function, $R$ is the autoregressive order, $\gamma_{i, r}$ is a time- and location-dependent autoregressive coefficient function, and $\omega_i$ is the innovation term, also varying over location $\bs{s}$ and time $t$. A similar model to \eqref{eq:ar_process} is considered in \cite{flumian2024stationary} in the context of stationary subspace analysis for time series.

We propose an identifiable autoregressive variational autoencoder (iVAEar) which extends the identifiability also to nonstationary autoregressive coefficients. In Section~2, we discuss iVAEar's model assumptions and identifiability conditions, and in Section~3 we introduce the iVAEar method to estimate the model. We demonstrate iVAEar's latent component estimation performance through comprehensive simulation studies in Section~4, and illustrate its multivariate spatio-temporal forecasting potential in Section~5. Finally, the paper is concluded in Section~6. All proofs are given in the supplement\footnote{https://github.com/mikasip/iVAEar} together with some additional material.

\section{Autoregressive latent component model and identifiability}

In this section, we introduce an autoregressive latent component model and its identifiability results under nonstationary data. We begin by establishing general identifiability conditions for autoregressive latent component models in Definition~\ref{def:identifiability_classes} and Theorems~\ref{theorem:1} and \ref{theorem:2}. We then examine specific cases that yield stronger identifiability results: first, we provide general results for the case where $R=0$ (Proposition~\ref{prop:component-wise}), followed by results for Gaussian latent components and Gaussian autoregressive latent components (Propositions~\ref{prop:gauss_identifiability} and \ref{prop:gaus_ar_identifiability}, respectively). Note, that although we focus on spatio-temporal data in the paper, all the results and estimation methods apply also for time series data by dropping the spatial location out of the equations.

In original iVAE \cite{Khemakhem2020}, the main assumption leading to identifiability of the latent component model is that an additional variable $\bs u \in \mathcal{U}$, where $\mathcal{U}$ is the domain of $\bs u$, is observed so that the latent components $\bs z$ have a conditional distribution $p(\bs z | \bs u) = \prod_{i=1}^P p(z_i| \bs u)$. In all previous iVAE methods, $\bs u$ has included information on temporal, spatial, or spatio-temporal location of the observation. In iVAEar, we assume that in addition to spatio-temporal location, we also have the previous $R$ observations in time, $\{\bs x(\bs s, t - 1), \dots, \bs x(\bs s, t - R)\} \coloneqq \bs x^-$, as the additional data. The autoregressive assumption leads to the following generative deep latent variable model:
\begin{align}
    p\left(\bs x, \bs z | \bs x^-; \bs u \right)  = p\left(\bs x|\bs z\right) p\left(\bs z | \bs z^-; \bs u\right),
    \label{eq:ar_generative_model}
\end{align}
where $\bs z^- = \{\bs z(\bs s, t - 1), \dots, \bs z(\bs s, t - R)\}$ is the set of previous latent components in time. Following \cite{Khemakhem2020}, the distribution $p(\bs x|\bs z)$ is defined as
\begin{align}
    p(\bs x|\bs z) = p_{\bs \epsilon}(\bs x - \bs f(\bs z)),
    \label{eq:x_distribution}
\end{align}
meaning that $\bs x$ decomposes into $\bs x = \bs f(\bs z) + \bs \epsilon$, where $\bs \epsilon$ is an independent noise vector. In non-noisy nonlinear BSS (\ref{eq:nonlinearBSS}), $p_{\bs \epsilon}$ can be modeled with a zero mean Gaussian distribution with infinitesimal variance. Further, it is assumed that the conditional latent distribution is part of the exponential family:
\begin{align}
    p_{\bs T, \bs \lambda}(\bs z| \bs z^-, \bs u)  = \prod_{i=1}^P \frac{Q_i(z_i,z_i^-)}{Z_i(\bs u)} \text{exp}\left[ \sum_{j=1}^k T_{i,j}(z_i,z_i^-) \lambda_{i,j}(\bs u) \right],
    \label{eq:exponential_family}
\end{align}
where $Q_i(z_i,z_i^-)$ is a base measure, $Z_i(\bs u)$ is a normalizing constant, 
$\bs T_i(z_i,z_i^-)=(T_{i,1}(z_i,z_i^-), \dots, T_{i,k}(z_i,z_i^-))^\top$  contains sufficient statistics, and $\bs \lambda_i(\bs u)=(\lambda_{i,1}(\bs u), \allowbreak \dots, \lambda_{i,k}(\bs u))^\top$ contains the parameters depending on $\bs u$. The dimension $k$ of each sufficient statistic $\bs T_i(z_i,z_i^-)$ and $\bs \lambda_i(\bs u)$ is assumed to be fixed. The formulation (\ref{eq:exponential_family}) reduces to general exponential family formula if the autoregressive order $R = 0$. The exponential family form in (\ref{eq:exponential_family}) includes variables $z_i$ generated through AR processes with any exponential family innovations if the location $\mu_i$ and AR coefficients $\gamma_i^r$ are constant. Some AR processes, such as processes with Gaussian or exponential distributed innovations, fall in this form even with nonstationary location and AR coefficients. The properties of Gaussian AR processes are discussed in more detail later in this section.

Assuming the generative model defined by the equations (\ref{eq:ar_generative_model})-(\ref{eq:exponential_family}), and nonlinear BSS (\ref{eq:nonlinearBSS}) problem, it is of interest to identify the latent components $\bs z$ as well as possible to obtain information about the true generative process behind the observed data. Hence, we next define two identifiability classes that can be obtained with sufficient assumptions. In following, we use the notation $\{\bs f^{-1}(\bs x(\bs s, t - 1)), \dots, \bs f^{-1}(\bs x(\bs s, t - R))\} \coloneqq \bs f^{-1}(\bs x^-)$ to denote the unmixing function applied to previous $R$ observations in time individually.

\begin{definition}
Consider the real parameter set $(\bs f, \bs T, \bs \lambda)$ and the estimated one $(\tilde{\bs f}, \tilde{\bs T}, \tilde{\bs \lambda})$ of mixing functions, sufficient statistics and natural parameters such that $p_{\bs f, \bs T, \bs \lambda}(\bs x | \bs x^-, \bs u) = p_{\tilde{\bs f}, \tilde{\bs T}, \tilde{\bs \lambda}}(\bs x | \bs x^-, \bs u)$ for all $\bs x, \bs x^- \in \mathcal{X}$ and $\bs u \in \mathcal{U}$. If there exists an invertible $Pk \times Pk$ matrix $\bs A$ and a vector $\bs c$ so that
\begin{align}
    \tilde{\bs T}(\tilde{\bs f}^{-1}(\bs x), \tilde{\bs f}^{-1}(\bs x^-)) = \bs A \bs T(\bs f^{-1}(\bs x), \bs f^{-1}(\bs x^-)) + \bs c
\end{align}
for all $\bs x, \bs x^- \in \mathcal{X}$, the set $(\bs f, \bs T, \bs \lambda)$ is identifiable up to an affine transformation. If $\bs A$ is a block permutation matrix, then the set $(\bs f, \bs T, \bs \lambda)$ is identifiable up to block-affine transformation.
\label{def:identifiability_classes}
\end{definition}

The block-affine identifiability is a stronger result, and often desirable. Block-affine identifiability is closely related to permutation and signed scale indeterminacy of $\bs z$ of linear BSS. To build intuition about how block-affine identifiability relates to the identifiability of the latent components $\bs{z}$, we next provide sufficient conditions on the sufficient statistics $\bs{T}$ in the case $R = 0$ that ensure identifiability of $\bs{z}$ up to permutation and component-wise nonlinearity.

\begin{proposition}
Assume that the set $(\bs f, \bs T, \bs \lambda)$ is identifiable up to block-affine transformation and that the autoregressive order $R=0$. Further assume: 
\begin{enumerate}[label=(\roman*)]
    \item A non-noisy BSS model (\ref{eq:nonlinearBSS}), i.e. that $\bs z = \bs f^{-1}(\bs x)$.
    \item There is a function $\tilde g_i: \mathbb{R}^k \rightarrow \mathbb{R}$ for all $i = 1,\dots P$ such that $\tilde g_i(\tilde{\bs T}_i(\tilde{z}_i)) = a_i\tilde{z}_i$, where $a_i \neq 0$.
\end{enumerate} Then we have that $\tilde{\bs f}^{-1}(\bs x) = \tilde{\bs z} = \bs P (g_1(z_1), \dots, g_P(z_P))^\top$, where $\bs P$ is a $P \times P$ permutation matrix and $g_1, \dots g_P$ are component-wise nonlinearities.
\label{prop:component-wise}
\end{proposition}

Assumption (ii) of Proposition~\ref{prop:component-wise} holds for most of common exponential family distributions such as Gaussian, beta, gamma, Pareto, Poisson and exponential distributions, which have sufficient statistic of the form $T(x) = x$ or $T(x) = \text{log}(x)$. If we have a noisy nonlinear BSS instead of non-noisy, there is an additional noise indeterminacy for each component. For the case $R > 0$ with autoregressive dependencies, similar results can be derived so that the component-wise nonlinearities would depend also on their previous values, i.e., that $\tilde{\bs f}^{-1}(\bs x) = \tilde{\bs z} = \bs P (g_1(z_1, z_1^-), \dots, g_P(z_P,z_P^-))^\top$. However, for specific autoregressive models, stronger identifiability results can be obtained. In particular, later in this section we demonstrate that for Gaussian autoregressive latent processes, the latent components can be identified up to permutation, location and scale transformations.

Next, we introduce two theorems that give sufficient conditions to achieve affine or block-affine identifiability. The main identifiability theorem is as follows:

\begin{theorem}
When the data are generated according the generative model in (\ref{eq:ar_generative_model})-(\ref{eq:exponential_family}), and the following holds:
\begin{enumerate}[label=(\roman*)]
    \item The set $\{\bs x \in \mathcal{X} | \rho_{\bs \epsilon}(\bs x) = 0\}$ has measure zero, where $\mathcal{X}$ is a domain of $\bs x$ and $\rho_{\bs \epsilon}$ is a characteristic function of the density $p_{\bs \epsilon}$ in (\ref{eq:x_distribution}).
    \item The mixing function $\bs f$ in (\ref{eq:x_distribution}) is injective.
    \item The sufficient statistics $T_{i,j}$ in (\ref{eq:exponential_family}) are differentiable with respect to $z_i$ almost everywhere, and the functions $T_{i,1}, \dots, T_{i,k}$ are linearly independent on any subset of $\mathcal{X}$ with positive measure.
    \item There exist $Pk + 1$ distinct points $\bs u_0, \dots, \bs u_{Pk}$ so that the $Pk \times Pk$ matrix $\bs L = (\bs \lambda(\bs u_1) - \lambda(\bs u_0), \dots, \lambda(\bs u_{Pk} - \lambda(\bs u_0))$ is invertible.
\end{enumerate}
Then, the set $(\bs f, \bs T, \bs \lambda)$ is identifiable up to affine transformation.
\label{theorem:1}
\end{theorem}
While the assumptions (i)-(iii) are not very restrictive, the assumption (iv) is crucial to understand as it restricts the identifiability only to cases where the parameters $\bs \lambda(\bs u)$ vary enough when $\bs u$ changes. Because of this assumption, the latent components are identifiable only when the exponential family parameters are nonstationary.

Although identifiability up to a affine transformation might already be useful, in most cases it is desirable to achieve block-affine identifiability. The next theorem gives sufficient conditions for such identifiability.

\begin{theorem}
    Assume that the assumptions of Theorem~\ref{theorem:1} hold. Further assume:
\begin{enumerate}[label=(\roman*)]
    \item The dimension of sufficient statistics is $k \geq 2$.
    \item The sufficient statistics $T_{i,j}$ are twice differentiable with respect to $z_i$.
    \item The mixing function $\bs f$ has all second-order cross derivatives.
\end{enumerate}
Then, the set $(\bs f, \bs T, \bs \lambda)$ is identifiable up to block-affine transformation.
\label{theorem:2}
\end{theorem}

Theorem~\ref{theorem:2}, combined with the additional conditions of Proposition~\ref{prop:component-wise}, essentially guarantees that latent components can be recovered up to permutation and component-wise nonlinearity. For example, Gaussian distributed latent components with unknown nonstationary mean and variance, with sufficient statistics $\bs T_i(z_i) = (z_i, z_i^2)^\top$, fall within Theorem~\ref{theorem:2}. In fact, we can show that for such Gaussian data the identifiability can be further reduced to permutation, scale and location shift, which is in par with identifiability results of linear BSS:

\begin{proposition}
\label{prop:gauss_identifiability}
Assume that the assumptions of Theorem~\ref{theorem:2} hold and that the data are generated through 
BSS model (\ref{eq:nonlinearBSS}). Further, assume that the latent components $z_i$ and the respective estimates $\tilde{z}_i$ are Gaussian, meaning that $\bs T_i(z_i) = (z_i, z_i^2)^\top$ and $\tilde{\bs T}_i(\tilde{z}_i) = (\tilde{z}_i, \tilde{z}_i^2)^\top$. Then we have that $\tilde{\bs z} = \bs P \bs \Lambda \bs z + \bs d$, where $\bs P$ is a permutation matrix and $\bs \Lambda$ is a diagonal matrix with non-zero diagonal elements.
\end{proposition}

Since our main focus in this paper is on Gaussian autoregressive latent components which always has $k \geq 2$, we refer the reader to \cite{Khemakhem2020} for $k = 1$ case, where sufficient conditions are provided for exponential family with $R = 0$. 
When the autoregressive process (\ref{eq:ar_generative_model}) is assumed for the latent components with Gaussian innovations, we have the following distribution:
\begin{align}
    &p(\bs z| \bs z^-, \bs u^t, \dots, \bs u^{t-R}) = \nonumber \\ &\prod_{i=1}^P \frac{1}{2\pi\sigma_i(\bs u^t)} \text{exp}\left[ \frac{\left(z_i - \mu_i(\bs u^t) - \sum_{r=1}^R( \gamma_{i,r}(\bs u^t)z^{t - r}_i - \mu_i(\bs u^{t-r}))\right)^2}{2 \sigma^2(\bs u^t)}\right],
    \label{eq:gaussian}
\end{align}
where $\bs u^t$ denotes the auxiliary variable for the observation $\bs x^t$.


\begin{proposition}
Assume that the assumptions of Theorem~\ref{theorem:2} hold and that the data are generated through BSS model (\ref{eq:nonlinearBSS}). Further assume that the latent components $z_i$ and the respective estimates $\tilde{z}_i$ are generated through the Gaussian AR process (\ref{eq:ar_process}) with $R \geq 1$. Then we have that $\tilde{\bs z} = \bs P \bs \Lambda \bs z + \bs d$, where $\bs P$ is a permutation matrix, $\bs \Lambda$ is a diagonal matrix with non-zero diagonal elements and $\bs d$ is a constant vector.
\label{prop:gaus_ar_identifiability}
\end{proposition}

Proposition~\ref{prop:gaus_ar_identifiability} gives the main identifiability conditions for the Gaussian autoregressive latent components. In practice, the conditions on the mixing function are not very restrictive. However, condition (iv) of Theorem~\ref{theorem:1} requires sufficient nonstationarity either in the AR coefficients $\gamma_{i,r}$ or in the variance $\sigma_i$. In Section~\ref{sec:ivae_ar}, we introduce an estimation method for estimating the generative model defined by equations (\ref{eq:ar_generative_model})-(\ref{eq:exponential_family}).



\section{Autoregressive identifiable variational autoencoder}
\label{sec:ivae_ar}

The iVAEar method is an autoregressive extension of spatio-temporal iVAE, introduced in \cite{sipila2024modelling}. It consists of an encoder $\bs g(\bs x, \bs u)$, a decoder $\bs h(\bs x)$ and an auxiliary function $\bs w(\bs u)$. As the true AR order $R$ is in general unknown, we use $W$ to refer to the AR order used in the iVAEar method. The method takes as an input the current observations $\bs x$ and their auxiliary data $\bs u$, and the $W$ previous observations in time and their auxiliary data $(\bs x^{t-r}, \bs u^{t-r})$, $r = 1, \dots, W$.

The encoder aims to estimate the unmixing function $\bs q$. It maps the observation and auxiliary data pair $(\bs x, \bs u)$ into the mean vector $\bs \mu_{\bs z| \bs x} \in \mathbb{R}^P$ and the variance vector $\bs \sigma_{\bs z| \bs x} \in \mathbb{R}^P$. For the current observation $\bs x$, the encoder's output is used for reparametrization trick \cite{kingma2013auto} to obtain a new latent representation $\bs z'$. The decoder aims then to estimate the mixing function $\bs f$ by trying to construct the original input $\bs x$ from $\bs z'$. For the previous observations $\bs x^{t-r}$, the encoder is used to obtain the corresponding latent component estimates $\bs u_{\bs z|\bs x, \bs u}^{t-r}$, which are provided by the mean function $\bs \mu_{\bs z| \bs x, \bs u}(\bs x^{t-r}, \bs u^{t-r})$. These are used to calculate the mean of the Gaussian latent distribution (\ref{eq:gaussian}). 

The auxiliary function $\bs w$ aims to estimate the function $\bs \lambda$ by mapping the auxiliary data $\bs u$ into parameters
$\bs \mu_{\bs z|\bs u}$, $\bs \sigma_{\bs z|\bs u}, \bs \gamma_{\bs z|\bs u}^{1}, \dots, \bs \gamma_{\bs z|\bs u}^{W}$, that estimate the true parameters of the autoregressive Gaussian distribution (\ref{eq:gaussian}). In addition, the auxiliary function is used to obtain the mean estimates $\bs \mu_{\bs z|\bs u}^{t-r}$ based on the auxiliary data $\bs u^{t-r}$ of the previous observations.

\begin{figure}
    \centering
    \includegraphics[width=0.7\linewidth]{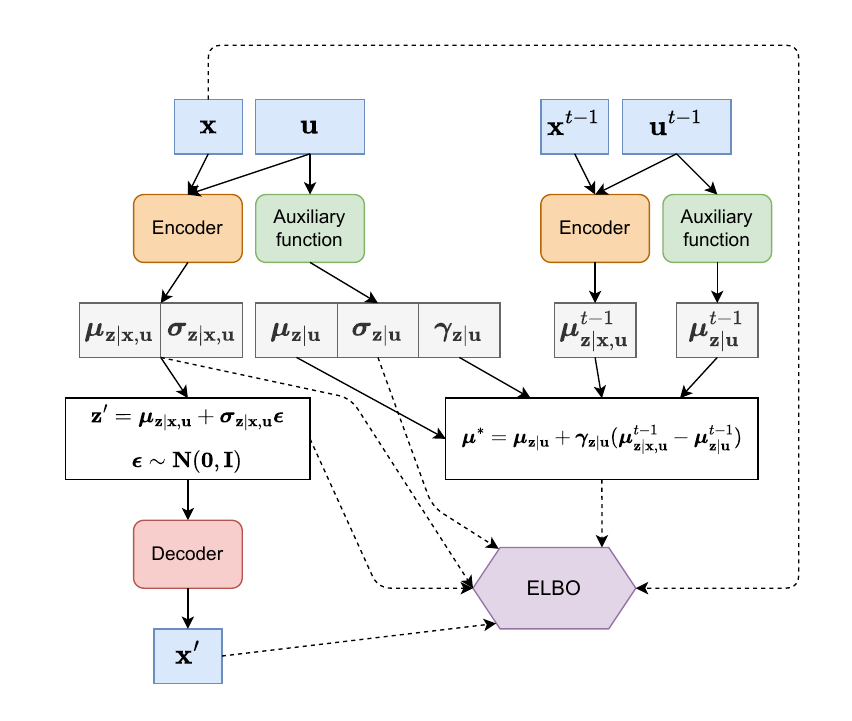}
    \caption{Schematic presentation of iVAEar method in $R=1$ case.}
    \label{fig:ivae_ar}
\end{figure}

The encoder, the decoder and the auxiliary function are modeled using deep neural networks with parameters $\bs \theta_{\bs g}, \bs \theta_{\bs h}, \bs \theta_{\bs w}$, that refer to the weights and biases of encoder, decoder and auxiliary function, respectively. The parameters $\bs \theta=(\bs \theta_{\bs g}, \bs \theta_{\bs h}, \bs \theta_{\bs u})^\top$ of the neural networks are optimized by minimizing the lower bound of the data log-likelihood, or evidence lower bound (ELBO):
\begin{align}
    \text{ELBO} = 
    E_{q_{\bs \theta}(\bs z|\bs x, \bs u)}\big (\text{log}\, p_{\bs \theta_{\bs h}}(\bs x | \bs z) + \text{log}\,p_{\bs \theta_{\bs w}}(\bs z | \bs z^-, \bs u) - \text{log}\,q_{\bs \theta_{\bs g}}(\bs z | \bs x, \bs u) \big),
\end{align}
where the first part, $p_{\bs \theta_{\bs h}}(\bs x | \bs z)$, controls the reconstruction accuracy and the second part, $\text{log}\,p_{\bs \theta_{\bs w}}(\bs z | \bs z^-, \bs u) - \text{log}\,q_{\bs \theta_{\bs g}}(\bs z | \bs x, \bs u)$, is the Kullback-Leibler divergence, which tries to keep the variational distribution $\text{log}\,q_{\bs \theta_{\bs g}} \allowbreak (\bs z | \bs x, \bs u)$ close to the prior distribution $\text{log}\,p_{\bs \theta_{\bs w}}(\bs z | \bs z^-, \bs u)$. Because Gaussian autoregressive latent data is assumed (\ref{eq:gaussian}), the distributions $p_{\bs \theta_{\bs w}}$, $q_{\bs \theta_{\bs g}}$ and $p_{\bs \theta_{\bs h}}$ are assumed Gaussian, ensuring that the estimated components follow the same distribution (\ref{eq:gaussian}). Specifically, we set $p_{\bs \theta_{\bs w}} = N(\bs z| \bs \mu^*, \text{diag}(\bs \sigma_{\bs z|\bs u}))$, where $\bs \mu^* = \bs \mu_{\bs z|\bs u} + \sum_{i=1}^R \gamma_{\bs z|\bs u}^{t-r}(\mu^{t-R}_{\bs z|\bs x, \bs u} - \mu^{t-R}_{\bs z|\bs u})$,  $q_{\bs \theta_{\bs g}} = N(\bs z| \bs \mu_{\bs z|\bs x, \bs u}, \text{diag}(\bs \sigma_{\bs z|\bs x, \bs u}))$ and $p_{\bs \theta_{\bs h}} = N(\bs x| \bs x', \beta \bs I)$, where $\beta > 0$ is a small constant that represents the variance of (\ref{eq:x_distribution}). By decreasing $\beta$, the weight of the reconstruction loss is increased in the loss function similarly as in $\beta$-VAE \cite{higgins2017betavae}. The whole iVAEar framework is illustrated in $R=1$ case in Figure~\ref{fig:ivae_ar}. For more details of iVAE framework, see \cite{Khemakhem2020, SipilaNonlinearSBSS, sipila2024modelling}.

For iVAEar, we construct the auxiliary data following \cite{sipila2024modelling} based on either spatial and temporal segmentation or spatial and temporal radial basis functions. In segmentation based algorithm, the spatial domain in divided into equally sized two dimensional square segments, and the temporal domain into equally sized one dimensional segments. The auxiliary variable then gives the spatial and temporal segments corresponding to the observation. In radial basis function based algorithm, multiple spatial and temporal node points are selected from spatial and temporal domains. The auxiliary variable, i.e. radial basis functions, are then constructed based on distance between the location of the observation and each of the node points. Segmentation based iVAEar is denoted iVAEar\_s and radial basis function based iVAEar is denoted iVAEar\_r in the rest of the paper. For further details of constructing the auxiliary data, see \cite{sipila2024modelling}.

If the underlying latent components satisfy the assumptions of Theorem 1 or Theorem 2, then we have the following consistency result.

\begin{theorem}
    Assume that the Theorem~\ref{theorem:1} or Theorem~~\ref{theorem:2} hold. Further assume that the family of the variational distributions $q_{\bs \theta_{\bs g}}(\bs z | \bs x, \bs x^-, \bs u)$ contains the distribution $p_{\tilde{\bs f}, \tilde{\bs T}, \tilde{\bs \lambda}}(\bs z | \bs x, \bs x^-, \bs u)$. Then iVAEar learns the true set $(\bs f, \bs T, \bs \lambda)$ up to the identifiability classes given by Theorems 1 and 2 in the limit of infinite data.
\end{theorem}

In AR Gaussian latent data case, when also $q_{\bs \theta_{\bs g}}$ is Gaussian, then by Proposition~3, iVAEar estimates the true latents $\bs z$ up to permutation, signed scale and location shift in the limit of infinite data. 

The auxiliary function of iVAEar enables the method to be used for spatio-temporal interpolation or forecasting purposes. Particularly, iVAEar\_r method provides smooth estimates of the spatio-temporal functions $\mu_i(\bs u^t)$, $\gamma_{i,r}(\bs u^t)$ and $\sigma_i(\bs u^t)$, $i = 1, \dots, P$, $r = 1, \dots, R$. These can be used to predict the latent components to new spatio-temporal locations, after which the predictions can be transformed into observation space by using the decoder of the trained iVAEar. The prediction capabilities of iVAEar are illustrated later in Section~5.


\section{Simulations}

The simulations of this paper are two-fold; in the first part in Section~\ref{sec:main_sim}, various simulations are performed under the assumption that the true AR order $R$ is known. 
In the second part in Section~\ref{sec:ar_sim}, the performance of iVAEar\_r is studied under the assumption that the true autoregressive order $R$ is unknown. The implementations of all iVAE and iVAEar variants together with the code to simulate the data in all considered settings and to reproduce the case study of Section~\ref{sec:case_study}, are available in GitHub\footnote{https://github.com/mikasip/NonlinearBSS}\textsuperscript{,}\footnote{https://github.com/mikasip/iVAEar}.

\subsection{Main simulations}
\label{sec:main_sim}
In this section, simulation studies are used to compare the performances of iVAEar\_r and iVAEar\_s against segmentation and radial basis function based spatio-temporal iVAE methods, iVAEs and iVAEr, respectively, as proposed in \cite{sipila2024modelling}, STBSS, and symmetric FastICA (FICA) with hyperbolic tangent nonlinearity \cite{hyvarinen1999fast}. In simulations, we 
generate the latent spatio-temporal fields $\bs z$ and a mixing function $\bs f$. 
We are particularly interested in performance in settings, where the variance and/or the AR coefficients of the latent fields $\bs z$ are varying in space and in time. Hence, we select one setting with nonstationary AR1 coefficient, one with nonstationary variance and one with both AR1 coefficient and variance nonstationary. In addition, each of the settings is considered with and without nonstationary spatio-temporal trend function. Next, we give all the simulation details and explain how 
$\bs z$ and 
$\bs f$ are generated.

In all simulations, we set the observed dimension $S=6$ and the latent dimension $P=6$. The number of spatial locations is $n_s=100$ and the number of time points is $n_t=500$. The spatial locations $\bs s_1, \dots \bs s_{n_s}$ are generated uniformly in the domain $[0, 1] \times [0, 1]$
, and the observations over time are set 
at times $t = 1, \dots, n_t$. The latent spatio-temporal fields are generated using the following vector AR process
. Assume the spatial field at time $t$ to be $\bs \delta(t) = (\delta(\bs s_1, t), \dots, \delta(\bs s_{n_s}, t))$. By using the vector AR process we have then
\begin{align}
    \bs \delta(t) = \sum_{r=1}^R \rho_r \bs K_r(t) \bs \delta(t - r) + \bs \epsilon_{\bs \delta}(t),
\end{align}
where $r=1,\dots,R$ is the order of AR process, $\rho_r$ is the baseline AR coefficient for the $r$th order, $\bs K_r(t)$ is a spatial kernel matrix for time $t$, which determines the temporal correlation with spatial locations, and $\bs \epsilon_{\bs \delta}(t)$ is a $n_s$-dimensional Gaussian noise vector with spatial covariance function $C(\epsilon_{\delta}(\bs s, t), \epsilon_{\delta}(\bs s', t))$, $s, s' \in \{ \bs s_1, \dots, \bs s_{n_s}\}$. If the kernel matrices $\bs K_r(t)$ are diagonal, the generated data have separable spatio-temporal covariance function, i.e., data do not have any spatio-temporal interaction. For the simulations, we set 
$R = 1$. As spatial covariance function for time $t$ we use variance modulated Matern covariance function
\begin{align}
    C(\epsilon_{\bs \delta}(\bs s, \bs s', t)) = \sigma(\bs s, t) \sigma(\bs s', t) \frac{1}{2^{\nu-1}\Gamma(\nu)}\left(\frac{|| \bs s  - \bs s' ||}{\phi}\right)^\nu K_\nu\left(\frac{|| \bs s  - \bs s' ||}{\phi}\right),
\end{align}
where $\sigma$ modulates the variance based on the time and spatial location, $K_\nu$ is a modified Bessel function of second kind, and $\phi$ and $\nu$ are range and shape parameters, respectively. The common Matern parameters for all settings are provided in the supplementary material. In the simulations, we consider data with and without trend. The spatio-temporal trend is generated as composition of cyclical and liner trends as follows:
\begin{align}
    \mu(s_1, s_2, t) = \theta_{s_1} s_1 + \theta_{s_2} s_2 + \theta_{t} t + \alpha \text{ sin}( \omega_{s_1} s_1 + \omega_{s_2} s_2 + \omega_{t} t + \omega_c).
    \label{eq:trend}
\end{align}
The parameters are generated so that $\theta_{s_1}, \theta_{s_2} \sim \text{Unif}(-3, 3)$, $\theta_t \sim \text{Unif}(-0.01, 0.01)$, $\omega_{s_1}, \omega_{s_2} \sim \text{Unif}(0.2, 4)$, $\omega_{t} \sim \text{Unif}(0.01, 0.1)$, $\omega_c \sim \text{Unif}(0, 2 \pi)$ and $\alpha \sim \text{Unif}(-2, 2)$.

\textbf{Setting 1.} The latent fields have constant variance $\sigma(\bs s, t) = 1$ and varying AR1 coefficients over space and time. The kernel matrix $\bs K_1(t)$ is a diagonal matrix with AR1 coefficients $\gamma(\bs s_1, t), \dots, \gamma(\bs s_{n_s}, t)$ in the diagonal for each spatial location $\bs s_1, \dots, \bs s_{n_s}$. The parameters $\gamma(\bs s, t)$ are generated as
\begin{align}
    \gamma(\bs s, t) = \text{cos}\left(\frac{2 \pi t b}{n_t} - c(\bs s)\right),
\label{eq:nonstat_ar}
\end{align}
where $b$ is a scale parameter and $c(\bs s)$ is a shift parameter. To obtain variability in space, we generate the shift parameters $c(\bs s)$ from the Gaussian distribution $N(0, 0.3)$ with Matern spatial covariance function with parameters $\phi_c, \nu_c$. The Matern parameters for shift are $\phi_{c_1}, \nu_{c_1}=(0.25, 5)$, $\phi_{c_2}, \nu_{c_2}=(0.15, 2)$, $\phi_{c_3}, \nu_{c_3}=(0.1, 3)$, $\phi_{c_4}, \nu_{c_4}=(0.3, 4)$, $\phi_{c_5}, \nu_{c_5}=(0.2, 1)$ for the latent components $z_1, \dots, z_5$. The scale parameters $b$ are generated from $\text{Unif}(1, 10)$ and the baseline AR1 parameters $\rho_r$ are generated from $\text{Unif}(0.6, 0.99)$ for each latent component.

\textbf{Setting 2.} The zero-mean latent fields $z^*_i$ are generated as in Setting 1. Then, the final latent fields are obtained as
$z_i(\bs s, t) = z^*_i(\bs s, t) + \mu_i(\bs s, t)$, where $\mu_i(\bs s, t)$ is generated as in (\ref{eq:trend}).

\textbf{Setting 3.} The latent fields have constant AR1 coefficients and varying variance over space and time. The kernel matrix is $\bs K_1(t)$ is identity matrix for all $t$. The spatial domain is divided randomly into 5 clusters and the time domain into 10 segments providing 50 spatio-temporal segments $S_1, \dots, S_{50}$, each having their own standard deviation $\sigma_1, \dots, \sigma_{50}$. The function $\sigma$ is then $\sigma(\bs s, t)=\sum_{k=1}^{50} \mathbbm{1}((\bs s, t) \in S_k) \sigma_k$, where $\mathbbm{1}$ is an indicator function giving 1, if the 
location $(\bs s, t)$ belongs in segment $S_k$, otherwise it gives 0. The baseline AR1 parameters $\rho_r$ are generated from $\text{Unif}(0.1, 0.9)$ for each latent component.

\textbf{Setting 4.} The zero-mean latent fields $z^*_i$ are generated as in Setting 3. Then, the final latent fields are obtained as
$z_i(\bs s, t) = z^*_i(\bs s, t) + \mu_i(\bs s, t)$, where $\mu_i(\bs s, t)$ is generated as in (\ref{eq:trend}).

\textbf{Setting 5.}
The latent fields have varying variances and varying AR1 coefficients over space and time. The fields are generated by combining settings 1 and 2. That is, we have an identical situation to Setting 2, but the function $\sigma$ is defined as in Setting 4.

\textbf{Setting 6.} The zero-mean latent fields $z^*_i$ are generated as in Setting 5. Then, the final latent fields are obtained as
$z_i(\bs s, t) = z^*_i(\bs s, t) + \mu_i(\bs s, t)$, where $\mu_i(\bs s, t)$ is generated as in (\ref{eq:trend}).

These simulation settings are considered to investigate how different types of nonstationarities affect the performance of the algorithms. The Settings 1 and 2 do not have any nonstationarity in variance, but do have nonstationary AR1 coefficient, meaning that the identifiability results hold for iVAEar methods, but not for iVAEs and iVAEr. In Settings 3-6 the variance is nonstationary, and hence the identifiability holds for all iVAE methods. Nonetheless, these settings are of interest when comparing performances when there are additional stationary or nonstationary autocorrelation present. Nonstationary trend is considered in Settings 2, 4 and 6 to see if that affects the performance. 


\textbf{Mixing function.} The observations $\bs x$ are obtained by applying a linear or nonlinear mixing function $\bs f_L$ to the generated latent components $\bs z$. The function $\bs f_L$ is generated using multilayer perceptron (MLP) following, e.g. \cite{Khemakhem2020, HyvarinenMorioka2016, Hyvarinen2019}. The parameter $L$ denotes the number of mixing layers used in MLP. Each layer $i$ consists of a $P \times P$ mixing matrix $\bs B_i$ and an activation function $\psi_i$. The matrices $\bs B_i$ are normalized to have unit length rows and colums in order to avoid vanishing of any of the latent components in the mixing process. The mixing function $\bs f_L$ can be then defined recursively as
\begin{align*}
    \bs f_L(\bs z) = \begin{cases}
        \psi_L(\bs B_L \bs z),\quad L = 1, \\
        \psi_L(\bs B_L \bs f_{L-1}(\bs z)),\quad L \in \{2,3,\dots\},
    \end{cases}
\end{align*}
where the activation function $\psi_L$ is linear for the first layer and exponential linear unit (ELU), given as
\begin{align*}
\psi_i(x)=\begin{cases}
    x,\quad x \geq 0, \\
    \text{exp}(x) - 1,\quad x < 0,
\end{cases}
\end{align*}
for the other layers. This results $\bs f_1$ with one layer being linear mixing, and when $L$ increases, the mixing function becomes increasingly nonlinear.

\textbf{Performance index.} The performance of the algorithms is measured using the mean correlation coefficient (MCC), which is 
also used for example in \cite{Hyvarinen2019, HalvaHyvarinen2020, SipilaNonlinearSBSS, sipila2024modelling}. MCC is a function of correlation matrix $\bs \Omega  = \text{Cor}(\bs z, \hat{\bs z})$ of the true and estimated latent components. MCC measures how similar the optimal permutation of $\bs \Omega$ is to $P \times P$ identity matrix, and is calculated as
\begin{align}
    \text{MCC}(\bs \Omega) = \frac{1}{P}\sup_{\bs P \in \mathcal{P}} \text{tr}(\bs P\, \text{abs}(\bs \Omega)),
\end{align}
where $\mathcal{P}$ is a set of all possible $P \times P$ permutation matrices, $\text{tr}(\cdot)$ is the trace of a matrix and $\text{abs}(\cdot)$ denotes taking elementwise absolute values of a matrix. The values of MCC vary in range $[0, 1]$, where $1$ is the optimal value, meaning that estimated components $\hat{\bs z}$ correlate perfectly with the true components $\bs z$.

\textbf{Model specifications.} All iVAE models have 3 hidden layers with 128 units in encoder, decoder and auxiliary functions. All hidden layers use leaky rectified unit (ReLU) activation function \cite{leaky_relu}. iVAEar\_r and iVAEr are set up with spatial resolution levels $H = (2, 9)$ and temporal resolution levels $G = (9, 17, 37)$. In iVAEar1\_s and iVAEs, $10 \times 10$ spatial segmentation is used by producing 100 equally sized segments, and temporal domain is divided into 100 segments, each of which contains 5 consecutive time points. For details of constructing the radial basis function based and segmentation based auxiliary variables, see \cite{sipila2024modelling}. All models are trained for 60 epochs with batch size of 64, and use the learning rate of 0.001 with polynomial decay of second-order over 10000
training steps, where the learning rate after the first 10000 training steps is 0.0001. STBSS uses two spatial ring kernels (0, 0.15) and (0.15, 0.3), and time lag of 1. These parameters were selected by training STBSS with multiple different parameters in each setting, and selecting the parameters that provided the best results on average. For more about STBSS and its parameters, see \cite{MuehlmannDeIacoNordhausen2022}.

\begin{figure*}
  \centering
  \includegraphics[width=1\textwidth]{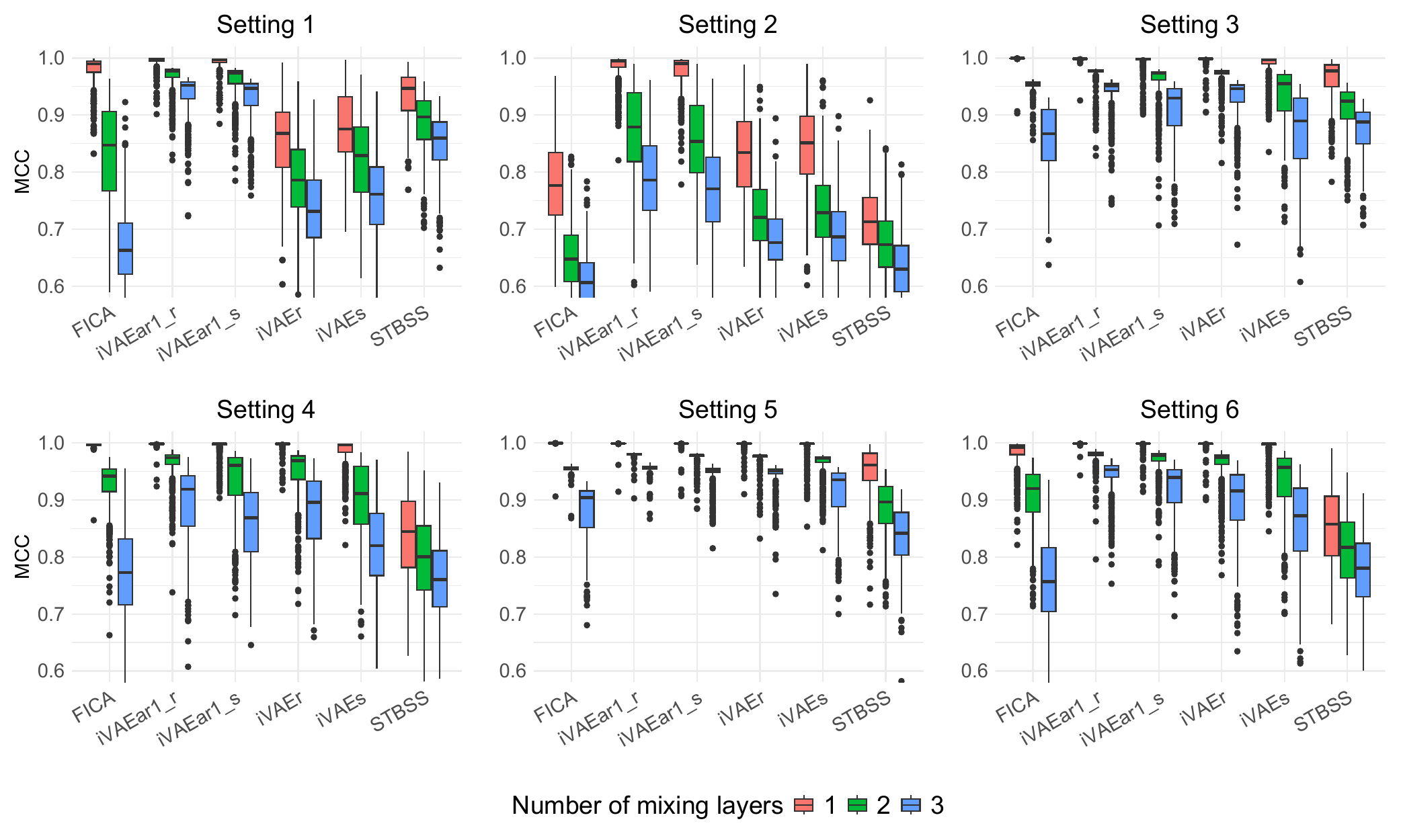}
  \caption{Mean correlation coefficients from 500 trials for Settings 1-6. The y-axis shows MCC (optimal value = 1), while the x-axis represents different methods. Box colors indicate the number of mixing layers in the mixing function.}
  \label{fig:sim1}
\end{figure*}

\textbf{Simulation results.} The results of the simulations are provided in Figure~\ref{fig:sim1}. Overall, the best results, especially in nonlinear scenarios, are obtained by iVAEar\_r, followed by iVAEar\_s in all settings. Nonstationary trend (Settings 2, 4 and 6) results in worse performance for all of the methods compared to settings where the trend is not present (Settings 1, 3 and 6). 

In Setting 1, where only AR1 coefficient is nonstationary, the latent components are successfully recovered only by iVAEar\_r and iVAEar\_s under nonlinear mixing. Under linear mixing, FICA performs nearly as well as iVAEar\_r and iVAEar\_s. STBSS is the fourth best performing method, followed by iVAEs and iVAEr. 

In Setting 2, where also nonstationary trend is added, iVAEar\_r and iVAEar\_s are the only methods with decent performance, although their performance also drops considerably in nonlinear settings. 

In Setting 3 with nonstationary variance, all of the methods perform relatively well. FICA and all iVAE based methods perform almost equally well under the linear mixing, but under the nonlinear mixing, FICA's performance suffers more. iVAEar based methods perform better than their iVAE counterparts, which is probably due to the fact that there are still stationary autocorrelation present in the latent components. 

In Setting 4, where the nonstationary trend is included into scenario of Setting 3, all of the methods lose performance. However, iVAEar\_r still maintains its performance nearly as well as in Setting 3, being clearly the best method. 

In Settings 5 and 6, where the variance and the AR1 coefficient are nonstationary, the results are very similar to the results of Settings 3 and 4, but the performances of FICA and iVAE methods are consistently slightly better due to the stronger nonstationarity. iVAE based methods maintain their performances better in nonlinear cases, and all of the methods perform slightly worse when the nonstationary trend is included.

Overall, autoregressive iVAE methods bring considerable improvement in performance as compared to the existing nonlinear STBSS methods. Based on the results, the methods can successfully estimate the latent components if there is either nonstationarity in autocorrelation or 
in variance. Nonstationary trend seems to be more challenging to tackle for the methods. Radial basis function based iVAEar, iVAEar\_r, is the best performing method in all of the settings, and is the recommended choice for nonlinear nonstationary STBSS problems.

\subsection{Sensitivity for AR order mismatch}
\label{sec:ar_sim}

In this section, we study how sensitive the best performing method, iVAEar\_r, is for AR order mismatch. The data are generated from Settings 1 and 5 with the true AR orders $R=1$ and $R=3$, and the latent components $\bs z$ are estimated using the iVAEar\_r with AR orders $W=1, 3, 5$, denoted iVAEar1\_r, iVAEar3\_r and iVAEar5\_r.

In $R=1$ scenario, the settings are identical to Settings 1 and 5 of the Section~\ref{sec:main_sim}. In $R=3$ scenario, the data are generated as in Settings 1 and 5, but the AR coefficients $\gamma_r(\bs s_i, t)$, $r = 1, \dots, R$, $i = 1, \dots, n_s$, are generated as in (\ref{eq:nonstat_ar}). The coefficients are then multiplied by constants $d_r$, where $d_r \sim \text{Unif}(0, 1)$, to create varying magnitudes to the components. The baseline AR coefficients are set to $\rho_r=1$, $r=1,\dots,R$. To guarantee the weak-sense stationarity of the AR process, defined in Definition~2 (supplementary material), the AR coefficients are scaled as follows:
\begin{align}
    \gamma_r(\bs s_i, t) = \frac{\gamma_r(\bs s_i, t)}{\text{max}_{i, t}( |\gamma_r(\bs s_i, t)| + |\gamma_r(\bs s_i, t)| + |\gamma_r(\bs s_i, t)|) + 0.01},
\end{align}
for each latent component $j=1,\dots, P$. This procedure guarantees $|\gamma_r(\bs s_i, t)| + |\gamma_r(\bs s_i, t)| + |\gamma_r(\bs s_i, t)| < 1$ for all $r=1, \dots, R$, $i = 1, \dots, n_s$, which is a sufficient condition for fulfilling the weak-sense stationarity.

\begin{figure*}[ht!]
  \centering
  \includegraphics[width=0.9\textwidth]{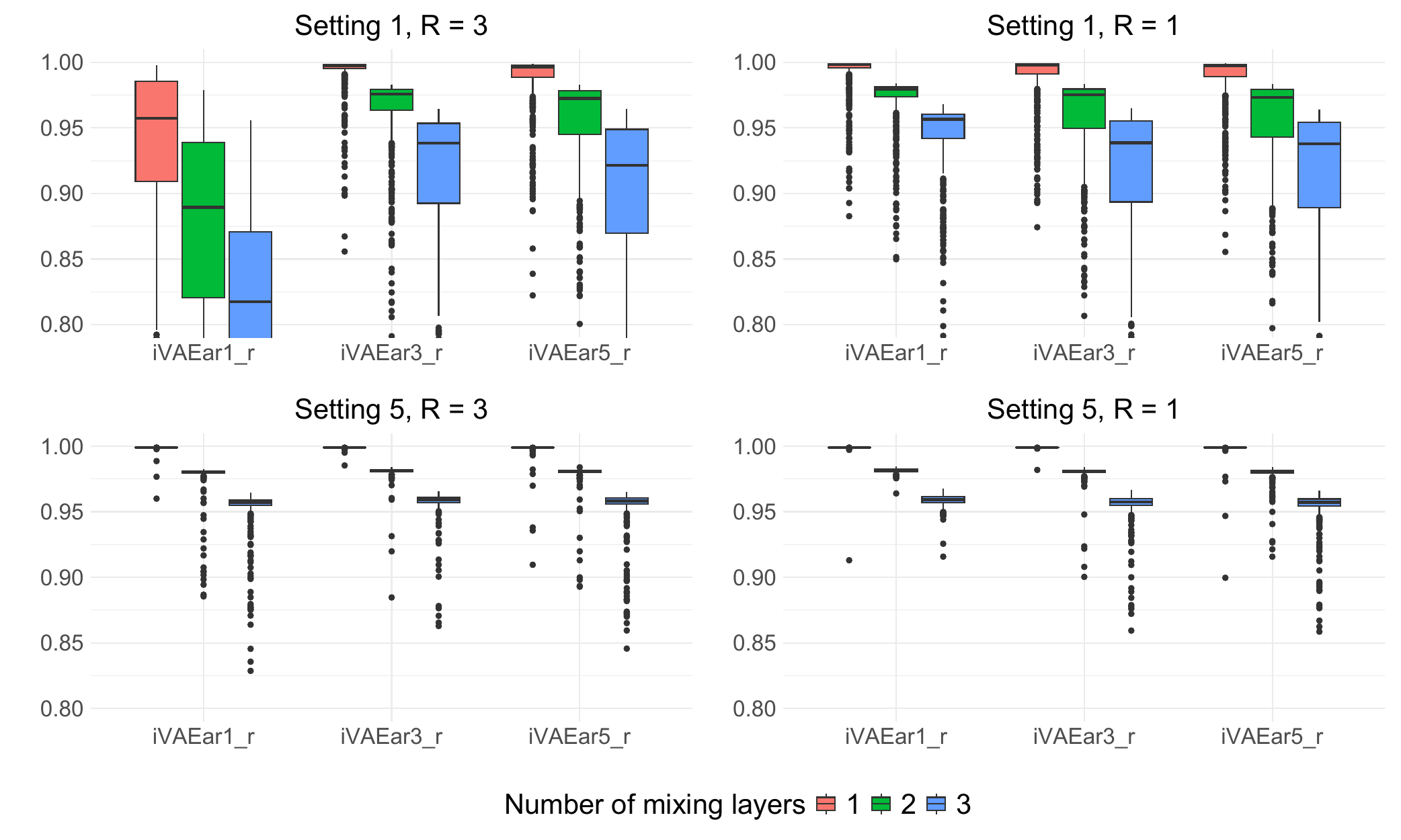}
  \caption{Mean correlation coefficients of 500 trials for Setting 1 (top) and Setting 5 (bottom) with $R=1$ and $R=3$. The y-axis shows MCC (optimal value = 1), while the x-axis represents different methods. Box colors indicate the number of mixing layers in the mixing function.}
  \label{fig:sim1_ar}
\end{figure*}

The results are presented in Figure~\ref{fig:sim1_ar}. In the case where only AR coefficients are nonstationary, the best performance is achieved when the true AR order $W=R$ is used in the model. Based on the results, it is safer to use larger $W$ as the performance drops only by little when $W>R$. The performance drops more significantly when too small $W$ is used in the model. In the case where also variance is nonstationary, the effect of incorrect AR order is negligible, although the correct AR order still produces the best performance. In general, based on the results, it is safer to use $W=3$ or $W=5$ in the model rather than $W=1$.

\section{Case study}
\label{sec:case_study}

We apply the iVAEar\_r and iVAEar\_s methods to an air pollution dataset \cite{airquality_datasets} to predict future values and compare their accuracy against iVAEr, spatio-temporal kriging \cite{cressie2011statistics}, ARIMA \cite{box2015time} and vector ARIMA (VARIMA) \cite{lutkepohl2005new}. Spatio-temporal kriging considers both spatial and temporal dependencies, while ARIMA models only temporal structures, making predictions separately for each station. Both kriging and ARIMA fit models univariately and do not account for cross-variable dependencies. In contrast, VARIMA models cross-dependencies between the variables through multivariate autoregressive process, but does modeling individually for each station. iVAEar\_r and iVAEr incorporate cross-variable dependencies through latent component decomposition and spatio-temporal trends. Additionally, iVAEar\_r estimates autoregressive structures of latent components for improved prediction.

The data consist of hourly air pollution and weather measurements from 64 stations in Athens, Greece, spanning 2020--2023. We use daily observations at 12 PM, resulting in $n_t = 1124$. The data include seven weather variables (wind speed U, wind speed V, dew point temperature, soil temperature, air temperature, relative humidity, precipitation) and four air pollution variables (PM10, PM2.5, NO2, O3). Precipitation is removed due to its predominantly zero values, yielding $S = 10$. Six stations lacking complete data are excluded, leaving $n_s = 58$. The remaining 162 missing observations are imputed using CUTOFF \cite{feng2014cutoff}. The last 24 time points serve as test data, while the first 1100 are used for training.

\begin{figure}
    \centering
    \includegraphics[width=0.5\linewidth]{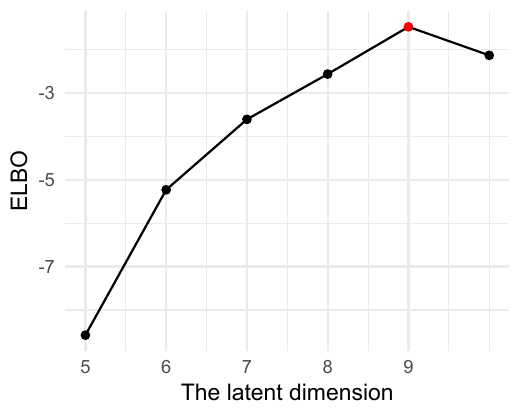}
    \caption{ELBO for different latent dimensions.}
    \label{fig:elbos}
\end{figure}

We estimate the latent dimension $P$ by fitting iVAEar\_r models with $P = 5, \dots, 10$, selecting the best model using knee-point detection and profile AIC (pAIC) \cite{sipila2024modelling}. ELBOs for different latent dimensions are shown in Figure~\ref{fig:elbos}. Both methods indicate $P=9$ as optimal, which is used in final models.

For forecasting with iVAEar\_r, iVAEar\_s and iVAEr, auxiliary data must remain within the training data bounds. Hence, we use seasonal periods $t_s = 1, \dots, 365$ instead of absolute time $t = 1, \dots, 1124$ and introduce a one-hot encoded year factor to allow inter-year variability. Spatial resolution levels are set to $H=(2,9)$, learning rate to 0.0001, variance parameter $\beta = 0.02$, batch size to 64, and training spans 40 epochs. A hyperparameter search optimizes temporal resolution for iVAEar\_r and iVAE\_r, segmentation sizes for iVAEar\_s, hidden units in the auxiliary function, and autoregressive order for iVAEar\_r and iVAEar\_s. The best parameters are selected by leaving 10 last time points of the training data for validation. Selected parameters are $G=(9, 17)$, $n_{\bs \theta_w}=16$, and $R=2$ for iVAEar\_r, $G=(9, 17)$, $n_{\bs \theta_w}=16$ for iVAE\_r and spatial segment size of 5000, temporal segment size of 5 and $R=3$ for iVAEar\_s.

For ARIMA, VARIMA and kriging, seasonal trends are removed as these methods assume seasonal stationarity. Seasonality is modeled as
\begin{align*}
    x_i(\bs s, t) = \beta_{0,i} + \beta_{1,i}\cos(2 \pi t / 365) + \beta_{2,i}\sin(2 \pi t / 365) + x_{res,i}(\bs s, t),
\end{align*}
where residuals $x_{res,i}$ are predicted using ARIMA and kriging. Kriging uses product-sum covariance models, while ARIMA selects the best model for each station via corrected AIC with AR orders $0, \dots, 5$, MA orders $0, \dots, 5$, and integration determined by the KPSS test \cite{kwiatkowski1992testing}. In VARIMA, we select the best model for each station based on AIC. The options are models with $\text{AR}=1, \dots, 8$ and $\text{MA}=0$, or a model with $\text{AR}=1$ and $\text{MA}=1$. Integration order was selected to be 1 for whole data, from options 1 or 0, based on better validation accuracy. VARIMA models with larger number of parameters caused numerical instability, and were hence not considered.

Prediction accuracy is measured using mean squared error (MSE):
\begin{align*}
MSE(\bs x_i, \hat{\bs x_i}) = \frac{1}{n}\sum_{j=1}^n(x_{i,j} - \hat{x_{i,j}})^2,
\end{align*}
where $\bs x_i$ contains true values and $\hat{\bs x_i}$ predicted ones. Combined accuracy is assessed via weighted MSE (wMSE):
\begin{align*}
wMSE(\bs X, \hat{\bs X}) = \frac{1}{S}\sum_{i = 1}^S \frac{MSE(\bs x_i, \hat{\bs x}_i)}{\sigma^2(\bs x_i)},
\end{align*}
where $\sigma^2(\bs x_i)$ is the variance of the deseasonalized variable.

Table~\ref{tab:forecast_results} presents forecasting results. iVAEar\_r outperforms competitors based on wMSE. VARIMA has the second best combined performance, and the best performance for predicting wind speeds. ARIMA has the second worst combined performance but excels for PM10, PM2.5, and NO2. Kriging and iVAEr perform similarly, with kriging being slightly better overall and excelling in soil temperature predictions. iVAEar\_r achieves the lowest errors for dew point temperature, air temperature and O3. iVAEar\_s has the best prediction performance for relative humidity, and has high accuracy for O3 as well, but its high errors on soil temperature, air temperature and NO2 makes it the worst method when considering the overall performance. Notably, O3 and relative humidity predictions benefit significantly from incorporating cross-variable dependencies, underscoring the advantage of iVAEar\_r, iVAEar\_s and VARIMA over univariate models. However, iVAEar\_s shows inconsistent performance in prediction and is suboptimal for this task. Its segmentation-based auxiliary variables lead to a highly non-continuous estimate of the trend function, which hinders the model's ability to generalize to future data. Therefore, iVAEar\_r is the preferred method for forecasting purposes. 


\begin{table}
\centering
\caption{Mean squared errors for predictions in time.}
\resizebox{1\linewidth}{!}{
\begin{tabular}{lrrrrrrrrrr|r}
  \hline
 & Wind Speed U & Wind Speed V & Dewpoint Temp & Soil Temp & Temp & Rel. Humidity & PM10 & PM2.5 & NO2 & O3 & wMSE \\ 
  \hline
iVAEar\_r & 1.57 & 6.16 & \textbf{3.42} & 1.08 & \textbf{3.44} & 64.15 & 81.31 & 30.42 & 106.87 & \textbf{93.09} & \textbf{0.49} \\
iVAEar\_s & 1.70 & 7.25 & 4.70 & 8.75 & 7.22 & \textbf{47.29} & 81.56 & 30.71 & 200.20 & 94.91 & 0.84 \\
iVAEr & 2.05 & 11.36 & 4.24 & 0.60 & 4.60 & 96.40 & 84.60 & 31.89 & 114.69 & 174.50 & 0.63 \\
  Kriging & 1.71 & 8.41 & 4.73 & \textbf{0.44} & 5.49 & 131.21 & 82.92 & 43.65 & 104.03 & 141.89 & 0.62\\ 
  ARIMA & 1.79 & 6.22 & 4.85 & 3.08 & 9.22 & 119.62 & \textbf{75.15} & \textbf{27.26} & \textbf{97.98} & 190.36 & 0.67 \\ 
  VARIMA & \textbf{1.54} & \textbf{5.75} & 4.11 & 1.64 & 8.29 & 65.68 & 75.70 & 35.80 & 99.06 & 121.81 & 0.56 \\
   \hline
\end{tabular}}
\label{tab:forecast_results}
\end{table}

\section{Conclusions and Discussion}

We have proposed a novel autoregressive iVAE method for nonlinear spatio-temporal BSS, extending identifiability results to cases with nonstationary autoregressive coefficients. Our simulation studies demonstrate superior latent component estimation compared to state-of-the-art methods, and real-world applications to air pollution and weather datasets show that iVAEar achieves significantly improved multivariate spatio-temporal prediction accuracy. Furthermore, we establish strong identifiability results, particularly for autoregressive Gaussian latent components.

A limitation of iVAEar is its reliance on a strict autoregressive assumption in time, making it optimal for separable spatio-temporal processes. Future work should explore extensions to nonseparable models and to more general graph structured data. As the identifiability under nonstationary AR coefficients was studied in this paper mainly for Gaussian innovations, the robustness of the method against innovations from other distributions should be studied in future.

In prediction tasks, careful hyperparameter selection and validation are necessary to prevent overfitting, and auxiliary variables must be chosen to ensure compatibility between training and test data. Additionally, iVAEar can be combined with univariate spatio-temporal prediction methods such as graphLSTM \cite{GaoLi2021}, allowing latent components to be predicted separately before reconstructing the observed data.

As iVAEar can be used for both time series and spatio-temporal data, it is a valuable method for latent component estimation and multivariate prediction across various fields, including environmental sciences, meteorology, and neuroscience, where applications often involve multiple temporal or spatio-temporal variables representing the same underlying phenomenon.

\begin{credits}
\subsubsection{\ackname}
We gratefully acknowledge support from the Vilho, Yrjö, and Kalle Väisälä Foundation (MS), the Research Council of Finland (grants 453691 to ST and 363261 to KN), and the HiTEc COST Action (CA21163) (KN, ST). We also thank CSC – IT Center for Science, Finland, for providing computational resources. Finally, we thank the three anonymous referees for their insightful feedback.

\subsubsection{\discintname}
The authors have no competing interests to declare that are
relevant to the content of this article. 
\end{credits}

\bibliographystyle{splncs04}
\bibliography{references}

\renewcommand\thesection{\Alph{section}}
\setcounter{section}{0}

\section{Lemmas for the autoregressive exponential families}

In this section, some useful Lemmas are given for univariate autoregressive exponential family distributions.

\begin{definition}[\textbf{Autoregressive models}] A generative model of $x$ is considered to be autoregressive, if it can be written as
\begin{align}
    x(\bs \theta^t) = \mu(\bs \theta^t) + \sum_{r=1}^R \gamma_{r}(\bs \theta^t) \Bigl( x^{t-r} - \mu(\bs \theta^{t-r}) \Bigr) + \omega(\bs \theta^{t}),
\end{align}
where $\bs \theta^t \in \mathbb{R}^m$ is the parameter vector at time step $t$, $\mu$ is a trend function, $\gamma_{1}, \dots, \gamma_{R}$ are the functions for autoregressive coefficients and $\omega$ is white noise so that $E(\omega(\bs \theta^t)) = 0$ and $\text{Var}(\omega(\bs \theta^t)) < \infty$ for all $t=1,\dots,T$, and $\text{Cov}(\omega(\bs \theta^t), \omega(\bs \theta^{t'})) = 0$ for all $t \neq t'$. To ensure local weak-sense stationarity for each $t$, the (complex) roots $y_i$ of the polynomial $1 - \sum_{i=1}^R \gamma_i(\theta_{t-i})y^i$ must satisfy $|y_i| > 1.$
\label{def:ar}
\end{definition}

\begin{definition}[\textbf{Autoregressive exponential family}] Assume an autoregressive model defined by Definition~\ref{def:ar}. The univariate distribution $p(x^t | \{x^{t-1:t-R}; \bs \theta^t\})$ belongs in univariate autoregressive exponential family, if its probability distribution can be written as
\begin{align}
    p(x^t | \{x^{t-1:t-R}; \bs \theta^t\}) = \frac{Q(x^t, \{x^{t-1:t-R}\})}{Z(\bs \theta^t)} e^{\sum_{j=1}^k T_j(x^t, \{x^{t-1:t-R}\})\lambda_j(\bs \theta^t)},
    \label{eq:ar_exponential_family}
\end{align}
where $Q$ is a base measure, $Z$ is a normalizing constant, $T_1, \dots, T_k$ are sufficient statistics and $\bs \theta^t$ is a parameter vector at time point $t$. The dimension $k \in \{1, 2, \dots \}$ is assumed to be minimal, meaning that the distribution $p$ cannot be written in form (\ref{eq:ar_exponential_family}) using a smaller $k' < k$.
\label{def:ar_exp}
\end{definition}

\begin{lemma}
Consider autoregressive exponential family distribution. The components of sufficient statistics $\bs T$ of the distribution are linearly independent. In other words, if there exists $\bs \alpha \in \mathbb{R}^k$ so that $\alpha_1 T_1(x^t, \{x^{t-1:t-R}\}) + \dots + \alpha_k T_k(x^t, \{x^{t-1:t-R}\}) = \bs 0$, then $\bs \alpha = \bs 0$.
\end{lemma}

\textit{Proof:} Assume that the components of $\bs T$ are not linearly independent. Then, there exists $\bs \alpha \in \mathbb{R}^k$, $\bs \alpha \neq \bs 0$, meaning that for some $i \in \{1, \dots, k\}$, $\alpha_i \neq 0$. By reordering the indices, we can assume that $\alpha_k \neq 0$. Then, we can write $T_k(x^t, \{x^{t-1:t-R}\}) = \sum_{j=1}^{k-1} \frac{\alpha_i}{\alpha_k} T_k(x^t, \{x^{t-1:t-R}\})$. Let $\lambda^*_j(\bs \theta^t) \coloneqq (\lambda_j(\bs \theta^t) + \frac{a_j}{a_k}\lambda_k(\bs \theta^t))$. Then, the term in the exponent of (\ref{eq:ar_exponential_family}) can be written as
\begin{align}    
\sum_{j=1}^k T_j(x^t, \{x^{t-1:t-R}\})\lambda_j(\bs \theta^t) &= \sum_{j=1}^{k-1} T_j(x^t, \{x^{t-1:t-R}\})\lambda_j(\bs \theta^t) + \sum_{j=1}^{k-1} \frac{\alpha_i}{\alpha_k} T_k(x^t, \{x^{t-1:t-R}\}) \\
&= \sum_{j=1}^{k-1} T_j(x^t, \{x^{t-1:t-R}\}) \left(\lambda_j(\bs \theta^t) + \frac{a_j}{a_k}\lambda_k(\bs \theta^t)\right)
\\
&= \sum_{j=1}^{k-1} T_j(x^t, \{x^{t-1:t-R}\}) \lambda^*_j(\bs \theta^t),
\end{align}
which contradicts the minimality of $k$ in Definition~\ref{def:ar_exp}. \qed

\begin{definition}[\textbf{Strongly exponential autoregressive distributions}] Exponential autoregressive distribution is considered strongly exponential if the following holds:
\begin{align}
    (\exists \, \bs \theta^t \in \mathbb{R}^m \, | \, \forall \, x^t, ..., x^{t-R} \in \mathcal{X}, \, \sum_{j=1}^k T_j(x^t, \{x^{t-1:t-R}\})\lambda_j(\bs \theta^t) = \text{const}) \implies l(\mathcal{X}) = 0 \text{ or } \bs \lambda(\bs \theta^t) = \bs 0,
\end{align}
where $l$ is a Lebesgue measure.
\label{def:strongly_ar_exp}
\end{definition}

Definition~\ref{def:strongly_ar_exp} says that strongly exponential distribution has the exponential component in its expression almost surely, and the distribution can be reduced only to base measure and normalizing constant on a set of measure zero.

\begin{lemma}
    Consider a strongly exponential autoregressive family distribution whose sufficient statistics $\bs T$ are differentiable almost everywhere. Then, $T'_j \neq 0$ for all $j = 1, \dots, k$ almost everywhere on $\mathbb{R}$.
\end{lemma}

\textit{Proof:} Assume that $p$ is strongly exponential autoregressive distribution. Let $\mathcal{X} = \cup_j\{x \in \mathbb{R}, T'_j(x) = 0\}$ and select any $\bs \theta$ for which $\bs \lambda(\bs \theta^t) \neq \bs 0$. Then, for all $x \in \mathcal{X}$, it holds that 
\begin{align}
    \sum_{j=1}^k T'_j(x^t, \{x^{t-1:t-R}\})\lambda_j(\bs \theta^t) &= 0 \\
    \implies \sum_{j=1}^k T_j(x^t, \{x^{t-1:t-R}\})\lambda_j(\bs \theta^t) &= \text{const}.
\end{align}
By Definition~\ref{def:strongly_ar_exp}, this means that $l(\mathcal{X}) = 0$. \qed

\begin{lemma}
    Consider a strongly exponential autoregressive family distribution of size $k \geq 2$ so that the sufficient statistics $\bs T$ are differentiable almost everywhere. Then, there exist $k$ distinct points $(x_1^t, \dots, x_1^{t-R}), \allowbreak \dots, (x_k^t, \dots, x_k^{t-R})$ such that the vectors $\bs T'(x_1^t, \{x_1^{t-1:t-R}\}), \dots, \bs T'(x_k^t, \{x_k^{t-1:t-R}\})$ are linearly independent in $\mathbb{R}^k$.
\end{lemma}

\textit{Proof:} Suppose that for any choice of such $k$ points, the vectors $\bs T'(x_1^t, \{x_1^{t-1:t-R}\}), \dots, \bs T'(x_k^t, \{x_k^{t-1:t-R}\})$ are not linearly independent, meaning that there are a subspace of $\mathbb{R}^k$ of dimension ar most $k - 1$ in which $\bs T'(\mathbb{R}^R)$ is included in. Thus, there exists $\bs \theta^t$ such that $\bs \lambda(\bs \theta) \in \mathbb{R}^k$ is a non-zero vector that is orthogonal to $\bs T'(\mathbb{R}^R)$. Because of the orthogonality, it holds for all $x_t,\dots, x^{t-R} \in \mathbb{R} $ that $\sum_{j=1}^k T'_j(x^t, \{x^{t-1:t-R}\})\lambda_j(\bs \theta^t) = 0$. By integrating, we find that $\sum_{j=1}^k T_j(x^t, \{x^{t-1:t-R}\})\lambda_j(\bs \theta^t) = \text{const}$. Since $\bs \lambda(\bs \theta^t)) \neq \bs 0$ and $l(\mathbb{R}) \neq 0$, the distribution cannot be strongly exponential, which contradicts the hypothesis.

\begin{lemma}
    Consider a strongly exponential autoregressive distribution of size $k \geq 2$ for which the sufficient statistics $\bs T$ are twice differentiable almost everywhere. Then it holds that
    \begin{align}
        \text{rank}\left( (T'_1(x^t, \{x^{t-1:t-R}\}), T''_1(x^t, \{x^{t-1:t-R}\})^\top, \dots, (T'_k(x^t, \{x^{t-1:t-R}\}), T''_k(x^t, \{x^{t-1:t-R}\})^\top\right) \geq 2 
        \label{eq:lemma4}
    \end{align}
    almost everywhere on $\mathbb{R}$.
    \label{lemma:4}
\end{lemma}

\textit{Proof:} Suppose there exists a set $\mathcal{X}$ so that $l(\mathcal{X}) > 0$, but the equation (\ref{eq:lemma4}) does not hold. In other words, for all $j \in \{1, \dots, k\}$ and $x \in \mathcal{X}$, the vectors $(T'_j(x^t, \{x^{t-1:t-R}\}), T''_j(x^t, \{x^{t-1:t-R}\})^\top$ are collinear. This means that there exists a vector $\bs \alpha \in \mathbb{R}^k$, $\bs \alpha \neq 0$, so that $\sum_{j=1}^k \alpha_j T'_j(x^t, \{x^{t-1:t-R}\}) = 0$. By integrating, we get $\sum_{j=1}^k \alpha_j T_j(x^t, \{x^{t-1:t-R}\}) = \text{const}$ for all $x \in \mathcal{X}$. Since $l(\mathcal{X}) > 0$, this contradicts the hypothesis.

\begin{lemma}
    Consider P strongly exponential autoregressive distributions of size $k \geq 2$ for which the sufficient statistics $\bs T_j$, $j = 1, \dots, P$ are twice differentiable almost everywhere. Let $\bs x \coloneqq (\bs x_1, \dots, x_P) \in \mathbb{R}^P$ and $\bs e^{(j,i)}(x_i) = (0, \dots, 0, T'_{j,i}(x_i), T''_{j,i}(x_i), 0, \dots, 0) \in \mathbb{R}^{2P}$, so that the non-zero entries are at indices $(2j, 2j + 1)$. Then the matrix $\bs E(\bs z) = (\bs e^{(1,1)}(x_1), \dots, \bs e^{(1,k)}(x_1), \dots, \bs e^{(P,1)}(x_P), \dots, \bs e^{(P,k)}(x_P)) \in \mathbb{R}^{2P \times Pk}$ has rank $2P$ almost everywhere on $\mathbb{R}^P$.
    \label{lemma:5}
\end{lemma}

\textit{Proof:} As the non-zero entries are at indices $(2j, 2j + 1)$, and there are $k$ columns in the matrix $\bs E$ for each $j=1,\dots,P$, the matrix $\bs E$ has at least the rank of $P$. By using Lemma~\ref{lemma:4}, it can be deduced that for each $j=1,\dots,P$, the submatrix $\bs E_j = (\bs e^{(j, 1)}(x_j), \dots, \bs e^{(j, k)}(x_j)$ has rank greater or equal to 2 almost everywhere on $\mathbb{R}$. Thus, it can be concluded that the rank of $\bs E$ is $2P$ almost everywhere on $\mathbb{R}^P$. \qed

\section{Proofs}

In this section, the proofs are provided for the main identifiability theorems and for all propositions. The proofs of Theorems 1 and 2 closely follow the approach of \cite{Khemakhem2020}, where the identifiability was proved for the exponential family without autoregressive structure.

\subsection{Proof of Proposition 1}
Since $(\bs f, \bs T, \bs \lambda)$ is identifiable up to block-affine transformation and $R=0$, we have
$\tilde{\bs T}(\tilde{\bs z}) = \bs A\bs T(\bs z) + \bs c$, where $\bs A$ is a block-permutation matrix and $\bs c$ is a constant vector.

Let $\pi$ be the permutation of $\{1,\dots,P\}$ induced by the block structure of $\bs A$. For each $i$, the $i$th block equation of the above is:
$\tilde{\bs T}_i(\tilde{z}_i) = \bs A_{i,\pi(i)}\bs T_{\pi(i)}(z_{\pi(i)}) + \bs c_i$
where $\bs A_{i,\pi(i)}$ is the $k \times k$ submatrix of $\bs A$ corresponding to the transformation from the $\pi(i)$th to the $i$th component, and $\bs c_i \in \mathbb{R}^k$ is the corresponding subvector of $\bs c$.

By applying $\tilde{g}_i$ to both sides for each $i$ and using assumption (ii), we have
$a_i\tilde{z}_i = \tilde{g}_i(\bs A_{i,\pi(i)}\bs T_{\pi(i)}(z_{\pi(i)}) + \bs c_i)$. Let $g_{\pi(i)}(z_{\pi(i)}) = \frac{1}{a_i}\tilde{g}_i(\bs A_{i,\pi(i)}\bs T_{\pi(i)}(z_{\pi(i)}) + \bs c_i)$. Then, we have
$\tilde{z}_i = g_{\pi(i)}(z_{\pi(i)})$.

The permutation $\pi$ defines a permutation matrix $\bs P$, giving us \\
$\tilde{\bs z} = \bs P (g_1(z_1), \dots, g_P(z_P))^\top$. \qed

\subsection{Proof of Theorem 1}
\textbf{Step 1.} Let us denote $\bs x^- = \{ \bs x^{t-1:t-R} \}$, $\bs x = \bs x^t$ and $\bs z =\bs z^t$. Suppose there are two sets of parameters $\bs \theta = (\bs f, \bs T, \bs \lambda)$ and $\tilde{\bs \theta} = (\tilde{\bs f}, \tilde{\bs T}, \tilde{\bs \lambda})$ such that $p_{\bs f, \bs T, \bs \lambda}(\bs x | \bs x^-, \bs u) = p_{\tilde{\bs f}, \tilde{\bs T}, \tilde{\bs \lambda}}(\bs x | \bs x^-, \bs u)$ for all $(\bs x | \bs x^-, \bs u)$. Then
\begin{alignat}{2}
&& \int_{\mathcal{Z}} p_{\bs f, \bs T, \bs \lambda}(\bs x, \bs z | \bs x^-, \bs u)d\bs z &= \int_{\mathcal{Z}} p_{\tilde{\bs f}, \tilde{\bs T}, \tilde{\bs \lambda}}(\bs x, \bs z | \bs x^-, \bs u) d\bs z \\
\implies&&  \int_{\mathcal{Z}} p_{\bs T, \bs \lambda}(\bs z | \bs x^-, \bs u)p_{\bs f}(\bs x | \bs z) d\bs z &= \int_{\mathcal{Z}} p_{\tilde{\bs T}, \tilde{\bs \lambda}}(\bs z | \bs x^-, \bs u)p_{\tilde{\bs f}}(\bs x | \bs z) d\bs z \\
\overset{(i)}{\implies}&&  \int_{\mathcal{Z}} p_{\bs T, \bs \lambda}(\bs z | \bs x^-, \bs u)p_{\bs \epsilon}(\bs x - \bs f(\bs z)) d\bs z &= \int_{\mathcal{Z}} p_{\tilde{\bs T}, \tilde{\bs \lambda}}(\bs z | \bs x^-, \bs u)p_{\bs \epsilon}(\bs x - \tilde{\bs f} (z)) d\bs z \\
\implies && \int_{\mathcal{X}} p_{\bs T, \bs \lambda}(\bs q(\bar{\bs x}) | \bs x^-, \bs u)p_{\bs \epsilon}(\bs x - \bs \bar{\bs x})|\text{det}(J_{\bs q}(\bar{\bs x}))| d\bar{\bs x} &= \int_{\mathcal{X}} p_{\tilde{\bs T}, \tilde{\bs \lambda}}(\tilde{\bs q}(\bs x) | \bs x^-, \bs u)p_{\bs \epsilon}(\bs x - \bar{\bs x}')|\text{det}(J_{\tilde{\bs q}}(\bar{\bs x}'))| d\bar{\bs x} \\
\implies&& \int_{\mathcal{\mathbb{R}^S}} \tilde{p}_{\bs T, \bs \lambda, \bs f, \bs u, \bs x^-}(\bs q(\bar{\bs x}))p_{\bs \epsilon}(\bs x - \bs \bar{\bs x}) d\bar{\bs x} &= \int_{\mathcal{\mathbb{R}^S}} \tilde{p}_{\tilde{\bs T}, \tilde{\bs \lambda}, \tilde{\bs f}, \bs u, \bs x^-}(\tilde{\bs q}(\bs x))p_{\bs \epsilon}(\bs x - \bar{\bs x}') d\bar{\bs x} \\
\implies&&  (\tilde{p}_{\bs T, \bs \lambda, \bs f, \bs u, \bs x^-} * p_{\bs \epsilon})(\bar{\bs x}) &= (\tilde{p}_{\tilde{\bs T}, \tilde{\bs \lambda}, \tilde{\bs f}, \bs u, \bs x^-} * p_{\bs \epsilon})(\bar{\bs x}') \\
\implies&& F[\tilde{p}_{\bs T, \bs \lambda, \bs f, \bs u, \bs x^-}](\omega) \varphi_{\bs \epsilon}(\omega) &= F[\tilde{p}_{\tilde{\bs T}, \tilde{\bs \lambda}, \tilde{\bs f}, \bs u, \bs x^-}](\omega) \varphi_{\bs \epsilon}(\omega) \\
\overset{(i)}{\implies}&& F[\tilde{p}_{\bs T, \bs \lambda, \bs f, \bs u, \bs x^-}](\omega) &= F[\tilde p_{\tilde{\bs T}, \tilde{\bs \lambda}, \tilde{\bs f}, \bs u, \bs x^-}](\omega) \\
\implies&& \tilde{p}_{\bs T, \bs \lambda, \bs f, \bs u, \bs x^-}(\bs x) &= \tilde p_{\tilde{\bs T}, \tilde{\bs \lambda}, \tilde{\bs f}, \bs u, \bs x^-}(\bs x)
\end{alignat}

\begin{itemize}
    \item In equation (26), J denotes Jacobian, a variable change $\bar{\bs x} = \bs f(\bs z)$ is introduced left hand side and $\bar{\bs x}' = \tilde{\bs f}(\bs z)$ to right hand side.
    \item In equation (27), $\tilde{p}_{\bs T, \bs \lambda, \bs f, \bs u, \bs x^-} = p_{\bs T, \bs \lambda}(\bs q(\bar{\bs x}^t) | \bs x^-, \bs u) |\text{det}(J_{\bs q}(\bar{\bs x}))| \mathbbm{1}_{\mathcal{X}}(\bs x)$ is introduced left hand side and similarly to right hand side. The indicator function $\mathbbm{1}_{\mathcal{X}}(\bs x)$ is defined as $\mathbbm{1}_{\mathcal{X}}(\bs x) = \begin{cases}
    1, \text{ when } \bs x \in \mathcal{X}, \\
    0, \text{ otherwise.}
    \end{cases}$
    \item In equation (28), $*$ denotes a convolution operator.
    \item In equation (29), $F$ denotes Fourier transform, and $\varphi_{\bs \epsilon} = F[p_{\bs \epsilon}]$.
    \item In equation (30), $\varphi_{\bs \epsilon}$ is dropped from both sides because of assumption (i) ($\varphi_{\bs \epsilon}$ is non-zero almost everywhere).
\end{itemize}
The step 1 guarantees that if the distributions with noise $\bs \epsilon$ are the same, then the noise-free distributions have to be the same.

\textbf{Step 2.} By starting from equation (31) and replacing the conditioning variable $\bs x^-$ with $\bs q(\bs x^-)=\{\bs q_x^{t-1}, \dots, \bs q_x^{t - R}\}$ (this can be done because $\bs f(\bs q(\bs x)) = \bs x$, meaning that $\bs q(\bs x)$ contains the same information as $\bs x$), denoting that the transformation $\bs q$ is applied to all $\bs x^i$, $i = t - 1, \dots, t - R$, we get the following form:
\begin{alignat}{2}
&& \tilde{p}_{\bs T, \bs \lambda, \bs f, \bs u, \bs x^-}(\bs x) &= \tilde p_{\tilde{\bs T}, \tilde{\bs \lambda}, \tilde{\bs f}, \bs u, \bs x^-}(\bs x) \\
\implies && p_{\bs T, \bs \lambda}(\bs q(\bs x) | \bs q(\bs x^-), \bs u)|\text{det}(J_{\bs q}(\bs x))| \mathbbm{1}_{\mathcal{X}}(\bs x) &= p_{\tilde{\bs T}, \tilde{\bs \lambda}}(\tilde{\bs q}(\bs x) | \bs q(\bs x^-), \bs u)|\text{det}(J_{\tilde{\bs q}}(\bs x))|\mathbbm{1}_{\mathcal{X}}(\bs x)
\end{alignat}
By taking a logarithm on both sides of equation (33) and replacing $p_{\bs T, \bs \lambda}$ and $p_{\tilde{\bs T}, \tilde{\bs \lambda}}$ with the form in equation (5), we get:
\begin{align}
    \text{log}|\text{det}(J_{\bs q}(\bs x))| + \sum_{i=1}^P (\text{log}Q_i(q_i(\bs x), q_i(\bs x^-)) - \text{log} Z_i(\bs u) + \sum_{j = 1}^k T_{i,j}(q_i(\bs x), q_i(\bs x^-))\lambda_{i,j}(\bs u)) = \nonumber \\
    \text{log}|\text{det}(J_{\tilde{\bs q}}(\bs x))| + \sum_{i=1}^P (\text{log}\tilde{Q}_i(\tilde{q}_i(\bs x), \tilde{q}_i(\bs x^-)) - \text{log} \tilde{Z}_i(\bs u) + \sum_{j = 1}^k \tilde{T}_{i,j}(\tilde{q}_i(\bs x), \tilde{q}_i(\bs x^-))\tilde{\lambda}_{i,j}(\bs u))
\end{align}

Let $\bs u_0, \dots, \bs u_{Pk}$ be the distinct points in assumption (iv). Then, we have $Pk + 1$ equations as in (34), one for each point. By subtracting the first equation from the others, for point $\bs u_l$, $l = 1, \dots, Pk$, we have
\begin{align}
    \sum_{i=1}^P \text{log} \frac{Z_i(\bs u_0)}{Z_i(\bs u_l)} + \sum_{i=1}^P\sum_{j = 1}^k ( T_{i,j}(q_i(\bs x), q_i(\bs x^-))(\lambda_{i,j}(\bs u_l) - \lambda_{i,j}(\bs u_0)) = \nonumber \\
    \sum_{i=1}^P \text{log} \frac{\tilde{Z}_i(\bs u_0)}{\tilde{Z}_i(\bs u_l)} + \sum_{i=1}^P\sum_{j = 1}^k ( \tilde{T}_{i,j}(\tilde{q}_i(\bs x), \tilde{q}_i(\bs x^-))(\tilde{\lambda}_{i,j}(\bs u_l) - \tilde{\lambda}_{i,j}(\bs u_0))
\end{align}
Let us define $\bar{\bs \lambda}(\bs u) = \bs \lambda(\bs u) - \bs \lambda(\bs u_0)$, and subtract $\sum_{i=1}^P \text{log} \frac{\tilde{Z}_i(\bs u_0)}{\tilde{Z}_i(\bs u_l)}$ both sides. Then we have
\begin{align}
    \sum_{i=1}^P\sum_{j = 1}^k ( T_{i,j}(q_i(\bs x), q_i(\bs x^-))(\bar{\lambda}_{i,j}(\bs u_l)) = 
    \sum_{i=1}^P \text{log} \frac{Z_i(\bs u_0)\tilde{Z}_i(\bs u_0)}{Z_i(\bs u_l)\tilde{Z}_i(\bs u_l)} + \sum_{i=1}^P\sum_{j = 1}^k ( \tilde{T}_{i,j}(\tilde{q}_i(\bs x), \tilde{q}_i(\bs x^-))(\bar{\tilde{\lambda}}_{i,j}(\bs u_l))
\end{align}

Let us write $b_l = \sum_{i=1}^P \text{log} \frac{Z_i(\bs u_0)\tilde{Z}_i(\bs u_0)}{Z_i(\bs u_l)\tilde{Z}_i(\bs u_l)}$ and set $\bs b = (b_1, \dots, b_{Pk})$. Let $\bs L$ be the matrix in assumption (iv), and $\tilde{\bs L}$ similar matrix for $\tilde{\bs \lambda}$. By expressing (36) in matrix form for all point $b_l$, $l=1,\dots,Pk$, we have:
\begin{alignat}{2}
    && \bs L^\top\bs T(\bs q(\bs x), \bs q(\bs x^-)) &= \tilde{\bs L}^\top \tilde{\bs T}(\tilde{\bs q}(\bs x), \bs q(\bs x^-)) + \bs b \\
    && \implies \bs T(\bs q(\bs x), \bs q(\bs x^-)) &= \bs (\bs L^\top)^{-1}\tilde{\bs L}^\top \tilde{\bs T}(\tilde{\bs q}(\bs x), \bs q(\bs x^-)) + (\bs L^\top)^{-1}\bs b \\ 
    && \implies \bs T(\bs q(\bs x), \bs q(\bs x^-)) &= \bs A \tilde{\bs T}(\tilde{\bs q}(\bs x), \bs q(\bs x^-)) + \bs c,
\end{alignat}
where $\bs A = (\bs L^\top)^{-1}\tilde{\bs L}^\top$ and $\bs c = (\bs L^\top)^{-1}\bs b$.
\newline
\textbf{Step 3.} By assumption (iii), Jacobian of $\bs T$ exists and is a $Pk \times P$ matrix of rank $P$. Because equation (39) holds, it also holds that $J(\bs T(\bs q(\bs x), \bs q(\bs x^-)) = \bs A J(\tilde{\bs T}(\tilde{\bs q}(\bs x), \tilde{\bs q}(\bs x^-)))$ and that $\text{rank}\Bigl(J(\bs T(\bs q(\bs x), \bs q(\bs x^-))\Bigr) = \text{rank}\Bigl(\bs A J(\tilde{\bs T}(\tilde{\bs q}(\bs x), \tilde{\bs q}(\bs x^-)))\Bigr)$. This leads to the fact that both $\bs A$ and $J(\tilde{\bs T}(\tilde{\bs q}(\bs x), \tilde{\bs q}(\bs x^-)))$ are of rank $P$.
\begin{itemize}
    \item If $k=1$, then A is invertible since it is a $P \times P$ matrix of rank $P$.
    \item If $k \geq 2$, define $\bar{\bs z} = \bs q(\bs x)$, $\bar{\bs z}^- = \bs q(\bs x^-)$ and $\bs T_i = (T_{i, 1}(\bar{z}_i, \bar{z}_i^-), \dots, T_{i, k}(\bar{z}_i, \bar{z}_i^-)$. Based on Lemma~3, it holds that for each $i=1,\dots, P$, there exists $k$ points $(\bar{z}_i^j, \bar{z}_i^{-,j})$, $j = 1, \dots, k$ such that $(\bs T'_i(\bar{z}_i^1, \bar{z}_i^{-,1}),  \allowbreak \dots, \bs T'_i(\bar{z}_i^k, \bar{z}_i^{-,k}))$ are linearly independent. Let us define $\bs Q = (J(\bs T(\bar{\bs z}^1, \bar{\bs z}^{-, 1})), \dots, J(\bs T(\bar{\bs z}^k, \bar{\bs z}^{-, k})))$, where each Jacobian is $Pk \times P$ matrix calculated with respect to $\bar{\bs z}^i$, and the vector $\bar{\bs z}^l$ and $\bar{\bs z}^{-, l}$ are defined as $\bar{\bs z}^l = (\bar{z}_1^l, \dots, \bar{z}_P^l)$ and $\bar{\bs z}^{-, l} = (\bar{z}_1^{-, l}, \dots, \bar{z}_P^{-, l})$. Similarly define matrix $\tilde{\bs Q}$ for Jacobians of $\tilde{\bs T}(\tilde{\bs q}(\bs f(\bar{\bs x}^l)), \tilde{\bs q}(\bs f(\bar{\bs x}^{-,l}))$ for the same points $l=1,\dots, k$. Then, by differentiating the equation (39) for each $\bs x_l$, we get the following in matrix form:
    \begin{align}
        \bs Q =\bs A \tilde{\bs Q}.
    \end{align}
    The matrix $\bs Q$ is invertible based on Lemma~3, and hence also $\bs A$ and $\tilde{\bs Q}$ are invertible. As we have invertible $\bs A$, the equation (39) says that the sufficient statistics are identifiable up to linear transformation and a constant. \qed
\end{itemize}

\subsection{Proof of Theorem 2} 
\textbf{Step 1.} The assumptions of theorem 1 holds, so we have
\begin{align}
    \bs T(\bs q(\bs x), \bs q(\bs x^-)) = \bs A \tilde{\bs T}(\tilde{\bs q}(\bs x), \tilde{\bs q}(\bs x^-)) + \bs c,
\end{align}
where $\bs c$ is a constant vector and $\bs A$ is an invertible $Pk \times Pk$ matrix. Let $(i, l, a, b)$ be four indices so that $1 \leq i \leq P$, $1 \leq l \leq k$ refer to the rows of the matrix $\bs A$, and $1 \leq a \leq P$, $1 \leq b \leq k$ refer to the columns of $\bs A$. Let $\bs v(\bs z) = \tilde{\bs q}(\bs f(\bs z)) : \mathcal Z \rightarrow \mathcal Z$. The function $\bs v$ is bijective as $\tilde{\bs f} : \mathcal Z \rightarrow \mathcal X$ and $\bs f: \mathcal Z \rightarrow \mathcal X$ are injective functions, and $\tilde{\bs q}(\tilde{\bs f}(\bs z)) = \bs z$.  Further, let there be two other indices $c, d \in \{1, \dots, P\}$, $c < d$ and denote $v_i^c = \frac{\partial v_i}{\partial v_c}$ and $v_i^{c,d} = \frac{\partial v_i}{\partial v_c\partial v_d}$. By differentiating (41) with respect to $z_c$, we get for each $1 \leq i \leq P$ and $1 \leq l \leq k$ the following:
\begin{align}
    \frac{\partial T_{i,l}(z_i, z^-_i)}{\partial z_c} = \sum_{a, b}  A_{i, l, a, b} \left ( \frac{\partial \tilde{T}_{a,b}(v_a(\bs z), v_a(\bs z^-))}{\partial v_a(\bs z)} \frac{\partial v_a(\bs z)}{\partial z_c} + \sum_{r=1}^R \frac{\partial \tilde{T}_{a,b}(v_a(\bs z), v_a(\bs z^-))}{\partial v_a(\bs z^{-r})} \frac{\partial v_a(\bs z^{-r})}{\partial z_c} \right ).
\end{align}
It holds that $\frac{\partial v_a(\bs z^{-r})}{\partial z_c} = 0$ for all $r=1,\dots, R$, as the values of previous time points do not depend on the value of current time point. Thus, we have
\begin{align}
    \frac{\partial T_{i,l}(z_i, z^-_i)}{\partial z_c} = \sum_{a, b}  A_{i, l, a, b} \left ( \frac{\tilde{T}_{a,b}}{\partial v_a(\bs z)} \frac{\partial v_a(\bs z)}{\partial z_c} \right ).
\end{align}
By differentiating (43) with respect to $z_d$, we get
\begin{align}
    0 = \sum_{a, b}  A_{i, l, a, b} \left( \frac{\partial \tilde{T}_{a,b}(v_a(\bs z), v_a(\bs z^-))}{\partial z_d} \frac{\partial v_a(\bs z)}{\partial z_c \partial z_d} + \frac{\partial \tilde{T}_{a,b}(v_a(\bs z))}{\partial^2 v_a(\bs z)}\frac{\partial v_a(\bs z)}{\partial z_c}\frac{\partial v_a(\bs z)}{\partial z_d} \right).
\end{align}

Let us define $\bs r_a^1(\bs z) = (v_a^{1,2}(\bs z), \dots, v_a^{P-1, P}) \in \mathbb{R}^{\frac{P(P-1)}{2}}$, $\bs r^2_a(\bs z) = (v_a^{1}(\bs z)v_a^2(\bs z), \dots, v_a^{P-1}(\bs z)v_a^{P}(\bs z)) \in \mathbb{R}^{\frac{P(P-1)}{2}}$, $\bs M(\bs z) = (\bs r^1_1(\bs z), \bs r^2_1(\bs z), \dots, \bs r^1_P(\bs z), \bs r^2_P(\bs z)) \in \mathbb{R}^{{\frac{P(P-1)}{2}} \times {\frac{P(P-1)}{2}}}$ and $\bs e^{(a,b)}(z_i) = (0, \dots, 0, T'_{a,b}(z_i), \allowbreak T''_{a,b}(z_i), 0, \dots, 0) \in \mathbb{R}^{2P}$, so that the non-zero entries are at indices $(2a, 2a + 1)$ and $\bs E(\bs z) = (\bs e^{(1,1)}(z_1), \allowbreak \dots, \bs e^{(1,k)}(z_1), \allowbreak \dots, \bs e^{(P,1)}(z_P), \allowbreak \dots, \bs e^{(P,k)}(z_P)) \in \mathbb{R}^{2P \times Pk}$. Finally, let $A_{i,l}$ be the $(i,l)$th row of the matrix $\bs A$. Then, by gathering the equation (43) for all pairs $(c, d)$, $c < d$ and pairs $(i, l)$ to a matrix form, we get
\begin{align}
    \bs M(\bs z)\bs E(\bs z) \bs A = \bs 0.
\end{align}

By Lemma~5, the matrix $\bs E$ is of rank $2P$ almost surely on $\mathcal{Z}$. Since the matrix $\bs A$ is full rank $Pk \times Pk$ matrix, we have $\text{rank}(\bs E \bs A) = 2P$ almost surely on $\mathcal{Z}$. Hence, by multiplying (45) from right with the pseudo-inverse of $(\bs E \bs A)$ we have
\begin{align}
    \bs M(\bs z) = \bs 0.
\end{align}
Particularly, $\bs r^2_a = \bs 0$ for all $a = 1, \dots, P$. This means that at each $\bs z \in \mathcal{Z}$, the Jacobian of $\bs v$, $J_{\bs v}$ has at most one non-zero entry in each row. Because $J_{\bs v}$ is invertible and continuous, the locations of the non-zero entries are fixed and do not change as function of $\bs z$. This proves that the function $\tilde{\bs q}(\bs f(\bs z))$ is a composition of a permutation and a point-wise nonlinearity.

\textbf{Step 2.} Without loss of generality, we assume that the permutation in $\bs v$ is identity. Let $\bar{\bs T}(\bs z) = \tilde{\bs T}(\bs v(\bs z)) + \bs A^{-1}\bs c$. In particular, $\bar{\bs T}$ is then a point-wise nonlinearity. Then, the equation (41) can be written as
\begin{align}
    \bs T(\bs z, \bs z^-) = \bs A \bar{\bs T}(\bs z, \bs z^-).
\end{align}
Let $\bs W = \bs A^{-1}$. Then, the equation (47) can be written for each component $1\leq i \leq P$ and sufficient statistic $1 \leq l \leq k$ as
\begin{align}
    \bar{T}_{i,l} = \sum_{a,b} D_{i, l, a, b} T_{a,b}(z_a, z_a^{-}).
\end{align}
By differentiating both sides with respect to $z_c$, $c \neq i$, we get
\begin{align}
    0 = \sum_b D_{i,l,c,b}\frac{\partial T_{c,b}'(z_a, z_a^-)}{\partial z_c}.
\end{align}
By Lemma~1, we know that $D_{i,l,c,b}=0$ for all $1 \leq b \leq k$, and since (49) holds for all $l$ and $c \neq i$, the matrix $\bs D$ must have a block diagonal form
\begin{align}
    \bs D = \begin{pmatrix}
\bs D_1 &  & \\
 & \ddots & \\
 & & \bs D_P
\end{pmatrix},
\end{align}
where each submatrix $D_1, \dots, D_P$ is a $k \times k$ matrix. Then, also the matrix $\bs A$ has the same block diagonal form, meaning that each submatrix $\bs A_i$ transforms $\bs T_i(\bs z, \bs z^-)$ into $\bar{\bs T}_i(\bs z, \bs z^-)$. Since $\bar{\bs T}$ is a point-wise nonlinearity, $\bs A$ has to be a permutation matrix. \qed

\subsection{Proof of Proposition 2} 
Based on the assumptions we have the following equalities 
\begin{align*}
    \tilde{z}_j = a_{11} z_i + a_{12} z_i^2 + c_1, \\
    \tilde{z}_j^2 = a_{21} z_i + a_{22} z_i^2 + c_2,
\end{align*}
for some constants $a_{11}, a_{12}, a_{21}, a_{22}, c_1$ and $c_2$. By squaring the first equation, we have $(a_{11} z_i + a_{12} z_i^2 + c_1)^2 = a_{21} z_i + a_{22} z_i^2 + c_2$. In order the equation to hold for all $z_i \in \mathcal{Z}$, it must hold that $a_{12} = 0$. Hence, we have that $\tilde{z}_j = a_{11} z_i + c_1$. \qed

\subsection{Proof of Proposition 3}
Since $(\bs f, \bs T, \bs \lambda)$ is identifiable up to block-affine transformation, we have
$\tilde{\bs T}(\tilde{\bs z}) = \bs A\bs T(\bs z) + \bs c$, where $\bs A$ is a block-permutation matrix and $\bs c$ is a constant vector.

Let $\pi$ be the permutation of $\{1,\dots,P\}$ induced by the block structure of $\bs A$, and $j = \pi_i$. Then we have that $\tilde{\bs T}_j(\tilde{z}_j^t, \dots, \tilde{z}_j^{t-R}) = \bs A_{i,j} \bs T_i(z_j^t, \dots, z_j^{t-R})$, where $\bs A_{i,j}$ is a $k \times k$ submatrix of $\bs A$ corresponding the indices $i$ and $j$. Because of Gaussian AR form (1), we have
\begin{align}
    &p(\bs z| \{ \bs z^{t-1:t-R} \}, \bs u^t, \dots, \bs u^{t-R}) = \nonumber \\ &\prod_{i=1}^P \frac{1}{2\pi\sigma_i(\bs u^t)} \text{exp}\left[ \frac{\left(z_i - \mu_i(\bs u^t) - \sum_{r=1}^R( \gamma_r(\bs u^t)z^{t - r}_i - \mu_i(\bs u^{t-r}))\right)^2}{2 \sigma^2(\bs u^t)}\right].
    \label{eq:gaussian}
\end{align}
and similar form for $\tilde{z}_j$ with parameter functions $\tilde{\mu}_j, \tilde{\sigma}_j, \tilde{\gamma}_{j,1}, \dots, \tilde{\gamma}_{j,R}$. Let $\gamma_{i,r} \coloneqq \gamma_{i, r}(\bs u^t)$, $\mu_{i,r} \coloneqq \mu_{i}(\bs u^t-r)$ and $\sigma_{i} \coloneqq \sigma_{i}(\bs u^t)$. By expanding the nominator in the exponential term, we have
\begin{align}
&(z_i^t)^2 - 2z_i^t\mu_{i,0} - 2z_i^t \sum_{r=1}^R \gamma_{i,r} z_i^{t-r} + 2 z_i^t \sum_{r=1}^R \gamma_{i,r} \mu_{i,r} + \mu_{i,0}^2 + 2 \mu_{i,0}\sum_{r=1}^R \gamma_{i,r}z_i^{t-r} + \nonumber \\ &(\sum_{r=1}^R \gamma_{i,r} z_i^{t-r})^2 - 2 (\sum_{r=1}^R \gamma_{i,r} z_i^{t-r})(\sum_{r=1}^R \gamma_{i,r} \mu_{i,r}) + (\sum_{r=1}^R \gamma_{i,r} \mu_{i,r})^2.
\end{align}

From this form, it is easy to see that the minimal sufficient statistics are $T_{i, 1} = (z^t_i)^2$, $T_{i, 2} = z^t_i$, $T_{i, 3, r} = z^t_iz_i^{t-r}$, $T_{i, 4, r} = z^{t-r}_i$, $T_{i, 5, r_1, r_2} = z^{t-r_1}_i z^{t-r_2}_i$, $r, r_1, r_2 \in \{1, \dots, R\}$. Similarly, we have the sufficient statistics $\tilde{\bs T_j}(\tilde{z}^t_j, \dots, \tilde{z}^{t-R}_j)$. Because of the block-affine identifiability, we have for each $k_1 \in \{1, \dots, k\}$ that
\begin{align}
    \tilde{T}_{k_1, j} = \sum_{k_2 = 1}^k a_{k_2, k_1, i} T_{k_2, i} + c_{i, k_1},
\end{align}
where $a_{k_1, k_2, i}$ and $c_{i, k_1}$ are constants. Importantly, we have for all $r = 0, \dots, R$ that $\tilde{z}_j^{t-r} = \sum_{k_2 = 1}^k a_{k_2, r_1, i} T_{k_2, i} + c_i$ and $(\tilde{z}_j^{t-r})^2 = \sum_{k_2 = 1}^k a_{k_2, r_2, i} T_{k_2, i} + c_i$. By squaring the first equation, we have that
\begin{align}
    (\sum_{k_2 = 1}^k a_{k_2, r_1, i} T_{k_2, i} + c_{i,r_1})^2 = \sum_{k_2 = 1}^k a_{k_2, r_2, i} T_{k_2, i} + c_{i, r_2}.
\end{align}
This equation holds only if the coefficients of the third order and above in the left hand side are zero, meaning that $a_{1, r_1, i}, a_{(3,r),  r_1, i}, a_{(5,r),  r_1, i} = 0$. Hence, we have for all $r_1= 0,\dots, R$ and $t = R + 1, \dots, T$ that
\begin{align}
    \tilde{z}_j^{t-r_1} = \sum_{r_2 = 0}^R b_{r_1, r_2, i} z_i^{t-r_2} + c_{r_1,i},
    \label{eq:ar_identifiability}
\end{align}
where $b_{r_1, r_2, i}$ are constants. Since (\ref{eq:ar_identifiability}) holds for all $t=R + 1, \dots, T$, we also have the following equations:
\begin{align}
    \tilde{z}_j^{t} &= \sum_{r = 0}^R b_{0, r, i} z_i^{t-r} + c_{0,i}, \nonumber \\
    \tilde{z}_j^{t} &= \sum_{r = 0}^R b_{R, r, i} z_i^{t+R-r} + c_{R,i},
    \label{eq:shifted_ar_identifiability}
\end{align}
where the second equation is obtained by shifting (\ref{eq:ar_identifiability}), for $r_1 = R$, $R$ time steps forward. From (\ref{eq:shifted_ar_identifiability}) we can deduce that all coefficients $b_{0, r, i}$, $r \neq 0$, have to be zero in order for the equations to hold for all $t \in \{R + 1, T\}$. Hence, we obtain $\tilde{z}^t_j = b_{0, 0, i} z_i^t + c_{0, i}$, which concludes the proof. \qed

\subsection{Proof of Theorem 3}

The lower bound of the data log likelihood (ELBO) (9) can also be written in the following format:
\begin{align}
    \text{ELBO} = 
    E_{q_{\bs \theta}(\bs z|\bs x, \bs u)}\big (\text{log}\, p_{\bs \theta}(\bs x | \bs x^-, \bs u) + \text{KL}(\text{log}\,q_{\bs \theta_{\bs g}}(\bs z | \bs x, \bs x^-, \bs u) || p_{\bs \theta}(\bs z | \bs x, \bs x^- , \bs u)) \big),
    \label{eq:elbo_alt}
\end{align}
where KL is Kullback-Leibler divergence and the set $(\tilde{\bs f}, \tilde{\bs T}, \tilde{\bs \lambda})$ are parametrized by $\bs \theta$. Minimizing ELBO given in (9) with respect to the parameters $(\bs \theta, \bs \theta_{\bs g})$ is equivalent to minimizing (\ref{eq:elbo_alt}), which means that in the limit of infinite data, the KL term eventually reaches zero, making the loss equal to the data log likelihood. Hence in this case, minimizing ELBO is equivalent to maximum likelihood estimation (MLE). As we assume that Theorem 1 or Theorem 2 hold, the consistency of MLE guarantees that the estimation converges to the corresponding identifiability class of the true set $(\bs f, \bs T, \bs \lambda)$ in the limit of infinite data. \qed

\section{Additional simulation details}

The parameters used in all simulation settings of Section 4.1, are provided in Table~\ref{tab:matern_params}.

\begin{table}[htb]
    \centering
    \caption{The parameters for the Matern covariance function in all simulation settings.}
    \begin{tabular}{lllllll}
    \toprule
         & IC1 & IC2 & IC3 & IC4 & IC5 & IC6 \\
             \midrule
         $\phi$ & 0.20 & 0.15 & 0.10 & 0.30 & 0.05 & 0.25 \\
         $\nu$ & 0.50 & 1.00 & 0.25 & 2.00 & 0.75 & 1.50 \\
         \bottomrule
    \end{tabular}
    \label{tab:matern_params}
\end{table}

\end{document}